\documentclass{article}

\usepackage{log_2026}
\usepackage[numbers,compress,sort]{natbib}
\usepackage{booktabs}       %
\usepackage{amsfonts}       %
\usepackage{nicefrac}       %

\usepackage{amssymb}
\usepackage{mathtools}
\usepackage{amsthm}
\usepackage{wrapfig}
\usepackage{array}         %
\usepackage{enumitem}      %
\setlist[itemize]{leftmargin=10pt, itemindent=0pt, topsep=1pt, itemsep=1pt, parsep=0pt}

\usepackage{xcolor}

\title[Why Freeze the Encoder?]{Why Freeze the Encoder? A Controlled Study of End-to-End Sequence–Graph Learning for User Behavior}

\theoremstyle{plain}
\newtheorem{theorem}{Theorem}[section]

\theoremstyle{definition}
\newtheorem{definition}[theorem]{Definition}

\theoremstyle{remark}

\usepackage{yc_package}
\AtEndPreamble{\RequirePackage[capitalize,noabbrev]{cleveref}}

\author[Y. Chen et al.]{%
  Yuen Chen$^{1}$\thanks{Corresponding authors: \texttt{yuenc2@illinois.edu}, \texttt{rizal.fathony@capitalone.com}.} \quad Yulun Wu$^{2}$ \quad Samuel Sharpe$^{2}$ \quad Igor Melnyk$^{2}$ \quad Nam H.~Nguyen$^{2}$ \\
  {\normalsize\bfseries Furong Huang$^{3}$ \quad C.~Bayan Bruss$^{2}$ \quad Rizal Fathony$^{2,*}$} \\[6pt]
  {\normalfont\small $^{1}$University of Illinois, Urbana-Champaign \quad $^{2}$Capital One \quad $^{3}$University of Maryland, College Park} \\
}

\begin{document}

\maketitle

\begin{abstract}
Modeling user behavior couples two distinct signals: a per-user sequence of events and a persistent user-to-user relation graph. Standard pipelines preserve these native formats by combining them in \emph{two stages}: a sequence encoder is pretrained, its per-user embeddings are frozen, and a graph model is trained on top to allow for cheap serving and cross-task reuse. We ask a simple question: does training the sequence and graph models \emph{end-to-end} help, and if so, when? To isolate the impact of the training strategy, we hold a standard sequence-encoder-to-GNN pipeline fixed and vary only the optimization method; on Brightkite, a matched control that starts both pipelines from the same pretrained encoder and changes only whether it is updated attributes a 10.6-point MRR gain to end-to-end updating. Across relationship prediction and fraud detection on five real-world datasets, we find that the benefit of end-to-end training is conditional: it improves clearly over frozen two-stage pipelines on relationship prediction, where the relation graph appears to carry much of the signal, whereas gains shrink where the predictive signal is largely node-local (fraud detection), a regime in which a text-fed relational graph transformer achieves competitive results. Furthermore, recasting the data as a temporal graph yields inconsistent results compared to this native sequence-graph formulation. Ultimately, this work provides a controlled empirical account of when jointly training the two structures helps.

\end{abstract}

\section{Introduction}\label{sec:intro}

\begin{figure}[t]
    \centering
\includegraphics[clip, trim=1mm 1mm 3mm 1mm, width=0.85\linewidth]{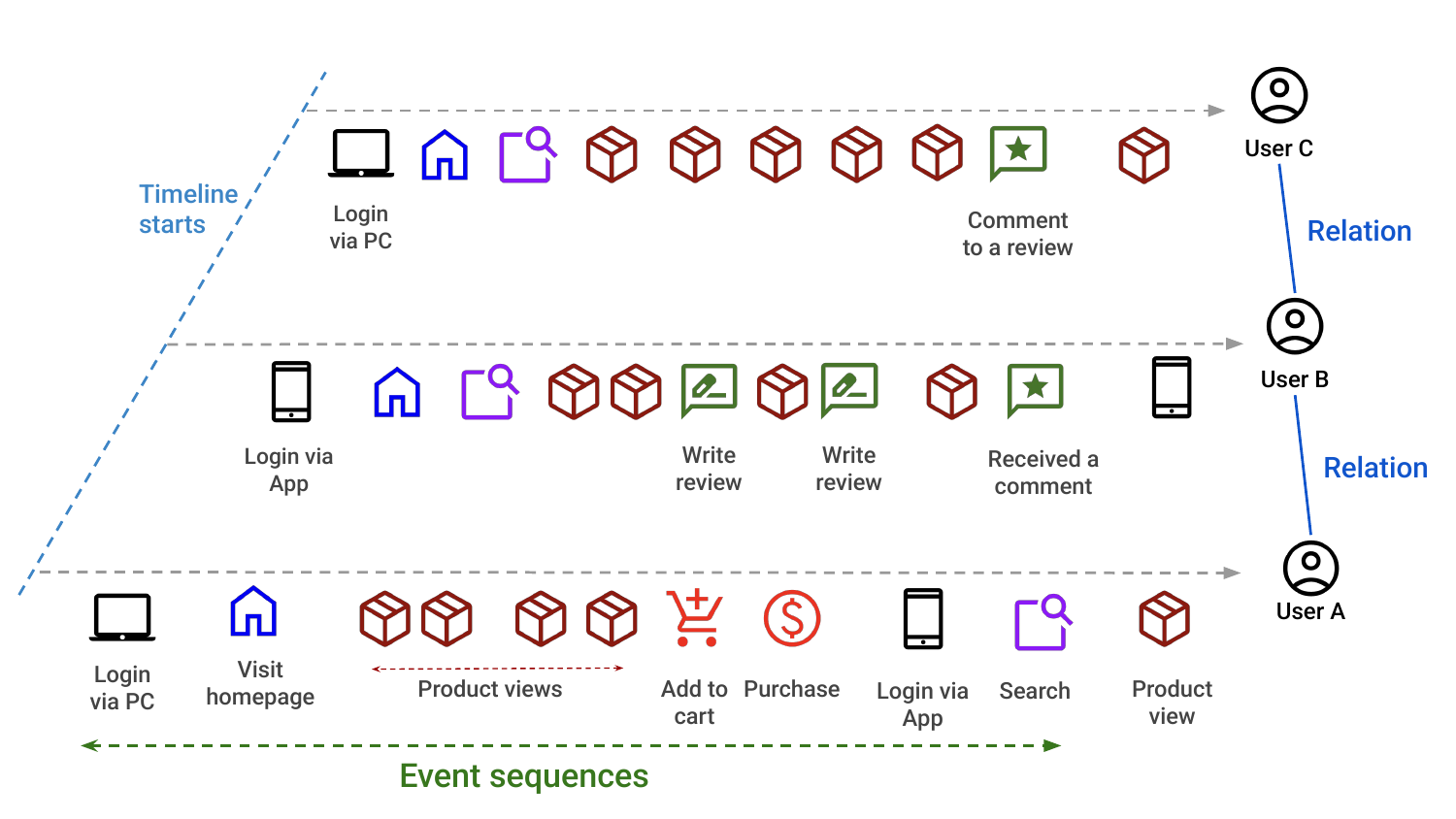}
\caption{A simple example of user behavior data in e-commerce. Each user has a sequence of events, while simultaneously also having a static user-to-user relationship such as friendship, family relation, sharing same address, sharing same devices, etc.}
    \label{fig:ecommerce}
\end{figure}

User behavior modeling has become a core component in many deployed industrial applications across a wide range of domains, including e-commerce \citep{zeng2019user,he2023survey}, social media \citep{abdelhafez2013survey, piao2018inferring, fang2023snapchat}, health \cite{li2020behrt,lee2023personalized,steinberg2023motor}, finance \citep{mashrur2020risk,ozbayoglu2020deep, hojaji2022behavioral,hernandez2024fraud}, cybersecurity \citep{lashkari2017survey, wang2017clickstream}, streaming platforms \citep{gayoavello2009session, amatriain2013netflix,covington2016youtube,zhao2023mbsrs}, and online gaming \cite{zeng2021human,chen2021sequential,yin2025modeling}. 
The behaviors of users inherently consist of sequential components, typically represented as an event sequence for each user, and relational components, usually represented as a relational graph. For example, in the case of e-commerce as illustrated in \cref{fig:ecommerce}, a user has an ordered event history, such as logins, page visits, check-ins, product views, reviews, payments, and many other events. Each user may also be connected to another user via user-to-user relationships, such as friendship, family relations, sharing same addresses, or sharing device relations.

The same structure recurs across domains: in healthcare, a patient's history of visits, measurements, and medications pairs with family or shared-environment links; in finance, a customer's logins, balance checks, and payments pair with joint-account, authorized-user, or shared-device links.

For modeling user behavior, each modality alone is incomplete: a purely sequential representation, as in personalized user event modeling \citep{letham2013sequential,boyd2020user} or sequential recommendation \citep{hidasi2015session,kang2018self}, omits user-to-user relationships, whereas forcing everything (including the sequential part) into a graph requires shoehorning each event into a node or edge. In practice, therefore, practitioners rarely pick just one modality; they \emph{combine} the two. They commonly do so in a decoupled, \emph{two-stage} fashion, however: a sequence encoder is trained first, its per-user embeddings are frozen, and a graph model is then trained on top of them \citep{fathonyintegrating}. The question at the center of this paper is whether it instead pays to train the two \emph{end-to-end}, and if so, when.

The difficulty of the graph route is representational, not only about training: many behavioral aspects do not map naturally onto nodes or edges \citep{fathonyintegrating}. Intransitive events (a login, check-in, or status update) involve no second entity, and persistent relations such as friendship or family carry no natural edge time. Yet the standard graph representations are either static or dynamic (temporal) graphs ~\citep{tgn_icml_grl2020,tgat_iclr20,trivedi2019dyrep}, the latter of which typically supports only event-based data.

Why is the two-stage pipeline the default, and why might end-to-end training not always help? Several reasons are practical rather than statistical. A frozen sequence encoder is trained once, and its embeddings are reused across many downstream models and refreshed on a slower schedule than the graph; end-to-end training couples the two and forfeits this reuse. Frozen embeddings are also cheap to precompute and serve, whereas joint training recomputes the sequence encoder over every sampled neighborhood. Finally, a frozen encoder can be pretrained on abundant unlabeled histories, while the joint model is bottlenecked by the (often scarce) task labels, so end-to-end training can overfit or destabilize when supervision is limited or a user's neighborhood carries little signal. These trade-offs, not an inability to connect the two modules, are why the modalities are often kept separate, and they make ``should we train end-to-end?'' a genuine, understudied decision.

To study this decision cleanly, we adopt a standard instantiation and use it as a testbed for a controlled study: a Transformer sequence encoder~\citep{vaswani_attention_2017,lim_temporal_2021} whose pooled output becomes the node feature of a standard static graph model~\citep{kipf2017semi,velickovic_graph_2018}, followed by a task-specific head, all trained end-to-end. Holding this pipeline fixed makes the training strategy the single variable of interest, letting us ask, with matched controls, when end-to-end optimization beats the two-stage default. We refer to this end-to-end instance as \ourmodel, since it bridges the sequence and graph models, and compare it against its two-stage counterparts as well as temporal-graph and relational-transformer reformulations of the same data.

Our contributions are:
\begin{itemize}
\item We frame a common but under-examined data regime (asynchronous per-user event histories together with a persistent, observed user-to-user relation) and make explicit when it is better served by a sequence--graph formulation than by temporal-graph or relational-transformer alternatives.
\item We set up controls that isolate the training strategy (graph-only and frozen two-stage baselines across five datasets and two tasks, relationship prediction and fraud detection; a no-relation ablation; and, on Brightkite, a matched freezing control in which two pipelines start from the same pretrained encoder and differ only in whether that encoder is frozen or fine-tuned).
\item We characterize \emph{when} end-to-end training helps: comparing it against the frozen two-stage baseline at matched architectures, we find that gains concentrate on relationship prediction, where the relation graph appears to carry much of the signal.
\end{itemize}
Our aim is a transferable lesson about when and why explicitly combining and jointly training the two structures helps.

\section{Related Work}\label{sec:related}

This section reviews related work on problem formulations for sequence modeling and graph modeling. We provide additional, more complete related work in Appendix \ref{sec:add_related_works}.

\textbf{User event sequence modeling}, such as personalized event sequence \citep{letham2013sequential,boyd2020user} and sequential recommendation \citep{hidasi2015session,kang2018self}, models a set of $N$ users $\mathcal{U} = \{u_1,\dots,u_N\}$, where each user maintains an independent sequence of events of length $M_i$, denoted as $S_i = [s_1^{(i)},\dots,s_{M_i}^{(i)}]$. Each item $s_j^{(i)}$ represents an event for user $u_i$, drawn from an event set $\mathcal{S} = \{1,\dots,|\mathcal{S}|\}$. This formulation, however, does not consider cross-user relationships, rendering an incomplete observation of user behaviors.

\textbf{Static graph modeling} \citep{kipf2017semi,velickovic_graph_2018}, on the other hand, directly encodes user-to-user relationship via its node and edge formulation $\mathcal{G} = (\mathcal{V}, \mathcal{E} )$. However, it cannot encode sequential information of user events when modeling user behaviors. 

\textbf{Temporal graph models} \citep{luo2022neighborhood,yu_towards_2023, pmlr-v235-gravina24a} represent the data as a sequence of timestamped edges $\mathcal{G} = [(u_1,v_1,t_1), (u_2,v_2,t_2), \dots]$, which does not natively accommodate a persistent static relation or node-local (intransitive) events \citep{nguyen2018continuous,kazemi2020representation,fathonyintegrating}. We defer a detailed account of when this formulation fits to \cref{sec:formulations} and an empirical diagnostic to \cref{sec:temporal_diag}.

\textbf{Two-stage sequence and graph integration}. 
The personal and relational event sequence (PRES) formulation by \citet{fathonyintegrating} augments the user event sequence formulation with relational information. Specifically, an event may be drawn from either a personal event set $\mathcal{P} = \{1,\dots,|\mathcal{P}|\}$ or a relational event set $\mathcal{R} = \{1,\dots,|\mathcal{R}|\}$. If an event is drawn from the relational event set, it must be accompanied by another user $v$. This formalization provides a way to combine sequential and relational information from the user event data. However, they also focused on event-based relationship information rather than static relational user-to-user information. 

\citet{fathonyintegrating} also proposed a method on integrating sequential and relational information that can be applied to our setting as well. Their methods uses sequence embeddings as graph node features. This validates the importance of both modalities, but its training pipeline separates sequence modeling from graph learning in a two-stage learning method. Our approach keeps the same practical motivation while using a single target-task objective to train the sequence and graph modules together.

\paragraph{Relational Deep Learning.}
Relational Deep Learning (RDL)~\citep{fey_relational_2023,Dwivedi2025RelationalDLAF} has recently emerged to natively model enterprise data structured across multi-table relational databases. Frameworks like RelBench~\citep{fey_relational_2023} represent databases as heterogeneous graphs defined by primary and foreign key links, while architectures such as RelGNN~\citep{chen2025relgnn} and the Relational Graph Transformer (RelGT)~\citep{dwivedi2025relgt} perform end-to-end message passing across these tables. While RDL shares our goal of end-to-end optimization over complex data, it diverges fundamentally in representation: RDL encodes user histories implicitly as unordered bags of child rows reached via $k$-hop sampling, whereas our setting treats the chronological, per-user sequence as a first-class, strictly ordered input. We detail this representational gap in \Cref{sec:formulations} and evaluate it empirically in \Cref{sec:exp_fraud}.

\textbf{Spatio-temporal graph models} \cite{yu_spatio-temporal_2018,wang_stgformer_2024} augment the static graph $\mathcal{G} = \{\mathcal{V}, \mathcal{E} \}$ with a time-synced feature sequence $[\mathbf{X}_1, \dots, \mathbf{X}_T]$, where $\mathbf{X}_t \in \mathbb{R}^{|\mathcal{V}| \times d}$, $T$ is the time duration, and $d$ is the feature dimension. The limitation is that the time dimension must be synced across all nodes. This fails for modeling user event sequences, as each user has independent events occurring at irregular intervals unsynced with other entities.

\textbf{Graph in sequential recommendation task.} There have also been efforts to augment the sequential recommendation task with graph models \cite{gao2020set,chang2021sequential,shui_sequence-graph_2022,zhang_dynamic_2023,baikalov2024end,cao2025graph}. Specifically, they define each sample as a sequence of items for a user, connected via sequential edges. They then add item-to-item collaborative edges denoting similarity, constructed by summarizing the user-item interaction matrix. While their formulation contains both sequence and graph components, it still takes only sequential data (sequences of items) as input and constructs edges within the sequence, rather than modeling cross-user relationships. Closest to our setting, \citet{baikalov2024end} learn graph and sequential representations end-to-end for recommendation, but over user--item interactions and with separate sequential and graph encoders whose outputs are fused late. We instead model within-user event sequences together with a persistent cross-user relation graph, stack the two so that the graph module operates directly on sequence-encoder outputs, and focus on the controlled question of whether that encoder should be frozen or trained jointly.

\section{Approach}

\subsection{Problem Formulation}\label{sec:setup}

Motivated by the problem of modeling user behaviors where each user carries sequential events and may have connections to another user via a static relationship, we define our formulation as follows:

\begin{definition}
\label{def:problem}
Let $\mathcal{U}=\{u_1,\dots,u_N\}$ be a set of users. Each user $u_i$ has an independent event history $S_i=[s_1^{(i)},\dots,s_{M_i}^{(i)}]$, where $s_j^{(i)}$ is the $j$-th event in that user's sequence; an event $s_j^{(i)}$ may be \emph{intransitive} (node-local), involving no other user and hence admitting no natural graph edge. Users are also connected by a static relation graph $G=(\mathcal{U},\mathcal{E})$, where an edge $(u_i,u_j)\in\mathcal{E}$ indicates a static user-to-user relation.
\end{definition}

This formulation combines the personalized event sequence formulation \cite{letham2013sequential,boyd2020user} with the static graph \citep{kipf2017semi,velickovic_graph_2018} representation, taking advantage of both modalities. It is also related to the PRES formulation~\citep{fathonyintegrating}, but differs in how relational information is represented: we adopt a static graph to reflect the persistent user-to-user relations in our setting, whereas PRES relies on an event-based relational structure. The downstream task in this setting could be relationship prediction (user-to-user link prediction), user classification, or any other user-level prediction problem.

\subsection{\ourmodel: An  End-to-End Instance}\label{sec:method}

\begin{figure*}[t]
    \centering
    \includegraphics[clip, trim=2mm 20mm 2mm 20mm, width=\linewidth]{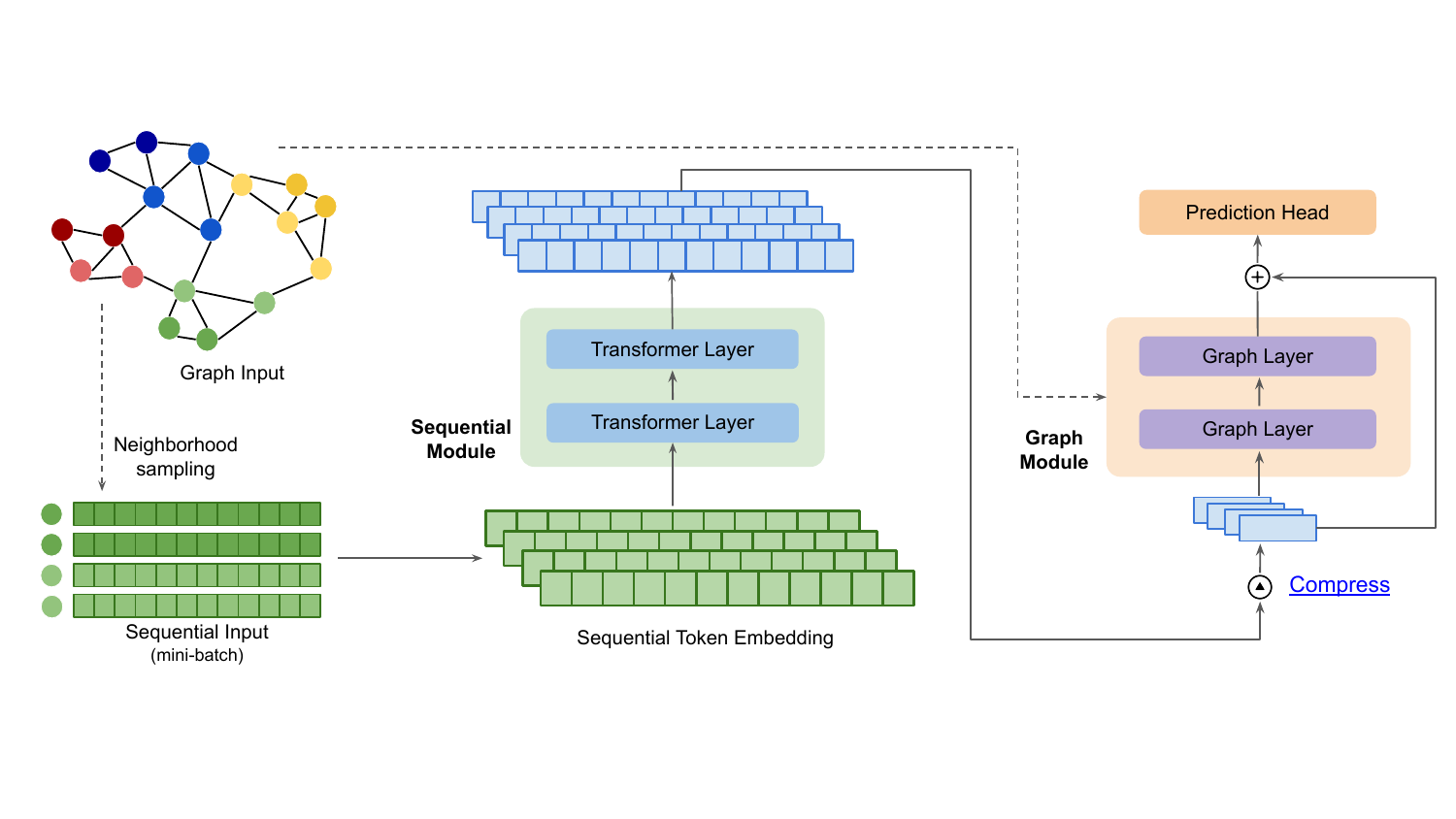}
    \caption{\ourmodel encodes each user's event history with a sequence module, compresses the sequence representation into user-level node features, propagates information over the static user-relation graph, and optimizes the full pipeline end-to-end for the downstream task.}
    \label{fig:overall_model}
\end{figure*}

As illustrated in \cref{fig:overall_model}, \ourmodel is a bidirectional Transformer sequence encoder whose mean-pooled output serves as the node feature of a standard GNN, followed by a task-specific prediction head. Every component is off-the-shelf; we study whether these three modules are optimized jointly end-to-end, or whether the encoder is pretrained and frozen before the GNN is trained on top.

For each user $i$ with input token embeddings $X_i\in\mathbb{R}^{M_i\times d}$, the forward pass is
\begin{align}
Z_i &= \mathrm{Enc}_\theta(X_i)\in\mathbb{R}^{M_i\times d},\\
z_i^{(0)} &= \mathrm{Compress}(Z_i)\in\mathbb{R}^{d_c},\\
z_i^{(\ell)} &= \mathrm{MP}^{(\ell)}\big(z_i^{(\ell-1)},\,\{z_j^{(\ell-1)}\}_{j\in\mathcal{N}(i)}\big),\quad \ell=1,\dots,L_g,\\
h_i &= z_i^{(L_g)}+z_i^{(0)},\qquad \hat{y}_i=\mathrm{Head}(h_i).
\end{align}
All parameters, the encoder $\theta$, the $L_g$ graph layers, and the prediction head, are trained jointly by minimizing a task loss $\sum_i \mathcal{L}(\hat{y}_i,y_i)$. In contrast, the two-stage baseline fixes $\mathrm{Enc}_\theta$ after pretraining, so no task-specific gradient reaches the sequence encoder. In our experiments, $\mathrm{Enc}$ is a bidirectional Transformer, $\mathrm{Compress}$ is mean pooling ($d_c=d$), and $\mathrm{MP}$ is a standard GCN, GAT, or TFConv~layer.

\section{Which Formulation Fits When?}\label{sec:formulations}

We separate two notions that are easily conflated: the \emph{problem setting} (the structure of the data and the downstream task) and the \emph{modeling formulation}, the primitives an architecture uses to represent it. (Concrete architectures such as TGN, RelGT, and our \ourmodel are instances of a formulation, not formulations themselves.) Our setting (Definition~\ref{def:problem}) couples per-user event histories with user-to-user relations, and this single setting can be cast into three dominant formulations: (i) \emph{temporal graphs}, which represent all information as a stream of timestamped edges; (ii) \emph{relational graph transformers} from Relational Deep Learning (RDL)~\citep{fey_relational_2023,Dwivedi2025RelationalDLAF}, which represent a multi-table database as a heterogeneous graph over primary/foreign-key links; and (iii) the \emph{sequence--graph} formulation we adopt, which keeps each user's event history as a first-class sequence and the user-to-user relation as a static graph. These formulations are not freely interchangeable: each is efficient in the setting it was designed for and pays an \emph{encoding tax} outside it, when data must be reshaped to fit primitives the formulation does not natively provide. We discuss the best- and worst-case setting of each below and summarize the trade-offs in \cref{tab:formulation_fit} (appendix).

\paragraph{Temporal graphs.}
Temporal (continuous-time dynamic) graph models~\citep{tgn_icml_grl2020,tgat_iclr20,trivedi2019dyrep} are the right tool when the predictive signal lives in \emph{interaction dynamics}. They excel when dyadic events, formalized as a stream of timestamped, optionally featured edges $(u,v,t,e)$, genuinely appear and disappear over time, and their timing, order, and recency carry the signal. They fit worst when two conditions of our setting hold. First, persistent relations (a friendship or family snapshot) lack a natural edge time. To run a temporal model we must fabricate one, drawing each static edge's timestamp uniformly at random from the overall span of observed event times; this manufactures a false ordering that these models are specifically built to exploit. Second, many user events are \emph{node-local and intransitive} (a login, a check-in, a status update) and have no second endpoint. Representing them requires artificial event-nodes that can become high-degree hubs. The diagnostic in \cref{sec:temporal_diag} is consistent with this account: on Brightkite and Gowalla, whose friendship edges are untimed and must be given synthetic timestamps, temporal methods do not reliably match the static graph baselines (\cref{tab:temporal_link_diag}). Where edge timestamps are instead real (the Amazon co-review streams), several temporal models become competitive. The failure is thus a property of the \emph{formulation mismatch} (fabricated time and forced event-nodes), not a weakness of the methods in their intended domain.

\paragraph{Relational graph transformers.}
RDL models such as RelGNN~\citep{chen2025relgnn} and the Relational Graph Transformer (RelGT)~\citep{dwivedi2025relgt} excel when the data is a \emph{rich, multi-table schema}: formally, a set of tables $\{T_k\}_{k=1}^K$ joined via primary/foreign-key references into a heterogeneous graph, where the signal lives in the join structure these keys induce. Their type-aware attention and composite message passing are designed for exactly this heterogeneity, and, importantly, they are trained end-to-end. Our setting sits near the opposite corner of that design space: a \emph{single} observed relation and a long, irregular per-user event \emph{sequence}. Encoded as an RDL graph, the sequence is not a first-class object; it becomes a set of rows reached via $k$-hop neighbor sampling and encoded as unordered typed tokens rather than an ordered sequence. Intransitive events, in turn, have no second-entity key to attach to. The model then spends capacity on relational heterogeneity that is largely absent here, while the ordered per-user history is encoded only implicitly, even though, on tasks like relationship prediction, that history carries much of the signal. We stress that RDL is \emph{applicable} here: a single-relation database is a legitimate, if degenerate, input. The distinction from our approach is therefore not ``end-to-end versus not'' (RDL is also end-to-end), but \emph{how the per-user history is represented}: a dedicated sequence encoder versus schema rows. We probe this representational gap where RelGT applies: Amazon fraud detection, cast as an RDL entity-classification task (\cref{tab:fraud_three}). There we find the gap \emph{negligible}: fraud is a node-local, low-prevalence regime in which a text-fed RelGT is competitive with our model. On relationship prediction, where the ordered per-user history and the observed relation carry the signal, we expect this representational gap to matter more. However, because RelGT's published scope excludes link prediction, we do not evaluate this representational gap there directly. Instead, our relationship-prediction experiments (\cref{sec:exp_friend}) isolate a different axis: the value of \emph{joint training} against matched frozen two-stage baselines.

\paragraph{Sequence--graph (our setting).}
The sequence--graph formulation fits best precisely where the other two struggle: each user has a rich individual event history that rewards a dedicated sequence encoder; the relation is persistent and observed rather than a stream; and events may be node-local. It reshapes nothing (sequences stay sequences and the static relation stays a static graph), so no synthetic time or artificial node is introduced. This is the regime of our relationship-prediction datasets, where explicitly combining the two structures improves over both graph-only and two-stage baselines (\cref{sec:exp_friend}). This formulation is a poor fit in three cases, which explicitly bound our claims: First, if the relation is genuinely \emph{dynamic} and its timing is the signal, a temporal graph is preferable. Second, if the ``graph'' is \emph{constructed} from the same item interactions the sequence already encodes (e.g., sequential recommendation), the relational and sequential views are partly redundant. Finally, if a true \emph{multi-table} schema is available, RDL is the more natural choice. A further, formulation-internal caveat is the \emph{compression bottleneck}: we summarize each history into a single node vector before message passing. This is cheap and decouples graph cost from sequence length, but it presumes that a per-user summary is what neighbors need; when fine-grained cross-user event alignment matters, this discards information.

\paragraph{A unifying principle.}
Across the three, the lesson is to \emph{match the formulation to where the signal and the structure actually live, and to fabricate nothing}. Time you do not have (synthetic edge timestamps), heterogeneity you do not have (a single relation dressed up as a database), and sequence structure you flatten (histories as schema rows) can each cost accuracy when the fabricated or discarded structure actually carries signal. The sequence--graph formulation is not universally superior; it is the efficient choice for the specific, common regime of rich per-user histories over a persistent single relation with node-local events, and we make that scope explicit rather than claiming generality.

\section{Experiments}\label{sec:exp}

We organize our experiments around user behavior modeling as described in Definition \ref{def:problem}. Specifically, we ask two main research questions: (1) does incorporating both users' event sequences and their static user-to-user graph information improve results over models that use only graph or sequence information?, and (2) does incorporating them in an end-to-end fashion further improve results over two-stage sequence-then-graph pipelines? We conducted these experiments on two different tasks: relationship prediction and fraud detection.

Additionally, we provide a separate section for diagnosing the performance of temporal graph methods when the problem formulation in Definition \ref{def:problem} is converted into a temporal graph formulation. As noted in the motivation of our paper, modeling user behavior may involve both dynamic event-based data and static information on user-to-user relations. \Cref{sec:temporal_diag} describes whether this conversion of static information into timestamped temporal graph events is a good fit.

\begin{table}[t]
\centering
\caption{Dataset statistics for relationship prediction (left) and Amazon fraud detection (right).}
\label{tab:dataset_stats}
\label{tab:fraud_data_stats}
\begin{minipage}[t]{0.49\textwidth}
\vspace{0pt}
\centering
\small
\setlength{\tabcolsep}{4pt}
\renewcommand{\arraystretch}{1.0}
\vspace{2pt}\textbf{Relationship prediction}\par\vspace{2pt}
\begin{tabular}{@{}lrrr@{}}
\toprule
Dataset & \# Users & \# Rel. & \# Events \\
\midrule
Brightkite    & 58,228  & 428,156   & 4,702,710 \\
Gowalla       & 196,591 & 1,900,654 & 6,442,289 \\
A-Movies      & 139,138 & 2,807,368 & 1,932,230 \\
A-Electronics & 254,064 & 657,050   & 2,281,128 \\
A-Clothing    & 185,986 & 18,078    & 1,573,869 \\
\bottomrule
\end{tabular}
\end{minipage}\hfill
\begin{minipage}[t]{0.49\textwidth}
\vspace{0pt}
\centering
\small
\setlength{\tabcolsep}{4pt}
\renewcommand{\arraystretch}{1.0}
\vspace{2pt}\textbf{Fraud detection}\par\vspace{2pt}
\begin{tabular}{@{}lrrr@{}}
\toprule
Statistics & Movies & Elec. & Cloth. \\
\midrule
\# Total Users & 139,138 & 254,064 & 185,986 \\
\# Labeled Users & 50,268 & 114,840 & 67,988 \\
\# Fraud (Pos.) & 13,912 & 7,456 & 1,658 \\
\# Non-Fraud (Neg.) & 36,356 & 107,384 & 66,330 \\
Fraud \% & 27.7 & 6.5 & 2.4 \\
\bottomrule
\end{tabular}
\end{minipage}
\vspace{0pt}
\end{table}

\subsection{Relationship Prediction}\label{sec:exp_friend}

\paragraph{Experimental Setup.}
We perform relationship prediction on the Brightkite and Gowalla datasets \citep{cho2011friendship,snapnets}, as well as Amazon reviews \citep{McAuley2015image}. Brightkite and Gowalla are location-based social networking platforms; each dataset contains location check-ins and friendship relationships among users. 
Each check-in, originally recorded in latitude and longitude, was converted into a Geohash-8 representation \citep{Niemeyer2008geohash}. Example geohashes include \texttt{9v6kpmr1}, \texttt{gcpwkeq6}, and \texttt{u0yhxgm1}, where nearby locations share common prefixes. 
For these datasets, each user possesses a sequence of geohash check-ins, which we further split into four tokens representing increasingly higher spatial resolutions. 
Both Brightkite and Gowalla treat friendship relations as static user-to-user properties without any timestamp information provided. 
For Amazon, we utilize the \textit{Clothing}, \textit{Electronics}, and \textit{Movies} categories. Here, each user's sequence consists of product IDs and ratings, and the static relation graph connects users who co-reviewed at least three products. Detailed data statistics are provided in \Cref{tab:dataset_stats}; ``A-'' denotes Amazon categories. 
For more info about the datasets, see \citet{fathonyintegrating}.
Positive relations are split into training, validation, and test sets. Only training relations form the message-passing graph; validation and test relations serve solely as prediction targets, and the sequence pretraining used by the two-stage baselines sees only users' event sequences, never relationship labels.

\paragraph{Main Compared Methods.}
As our task contains both event sequence and static relational information, our main comparable baselines are models that can operate on the static graph formulation.

\begin{enumerate}
    \item \textit{\textbf{Graph-Only Models:}} GCN~\citep{kipf2017semi}, GAT~\citep{velickovic_graph_2018}, and TFConv~\citep{ijcai2021p214} operate on the static user-to-user relation graph, ignoring each user's individual event sequences.

    \item \textit{\textbf{Two-Stage Training:}} Following \citet{fathonyintegrating}, GCN+S, GAT+S, and TFConv+S extend static graphs by incorporating sequence embeddings as node features. These embeddings are pre-trained using a bidirectional Transformer encoder via masked token prediction on user sequence tokens and remain frozen during graph training.

    \item \textbf{\textit{\ourmodel:}} Our model stacks a bidirectional Transformer sequence model with various GNN backbones, trained jointly in an end-to-end manner. Unlike the two-stage training approach, \ourmodel does not utilize frozen embeddings; instead, it optimizes the Transformer and GNN components simultaneously for the target task. 

\end{enumerate}

\begin{table*}[t]
\caption{Relationship prediction performance across five datasets (higher is better). The main comparison uses methods aligned with the static-relation formulation: graph-only baselines, frozen two-stage sequence+graph baselines, and end-to-end \ourmodel variants. Best results are \textbf{bolded} and the runner-ups are \underline{underlined}.}
\centering
\small
\setlength{\tabcolsep}{4pt}
\label{tab:link_pred_full}
\resizebox{0.98\textwidth}{!}{%
\begin{tabular}{@{}l*{10}{@{\hspace{4pt}}c}@{}}
\toprule
& \multicolumn{2}{c}{\textbf{Brightkite}} & \multicolumn{2}{c}{\textbf{Gowalla}} & \multicolumn{2}{c}{\textbf{Movies}} & \multicolumn{2}{c}{\textbf{Electronics}} & \multicolumn{2}{c}{\textbf{Clothing}}\\
\cmidrule(lr){2-3}\cmidrule(lr){4-5}\cmidrule(lr){6-7}\cmidrule(lr){8-9}\cmidrule(lr){10-11}
\textbf{Method} & \textbf{MRR} & \textbf{H@5} & \textbf{MRR} & \textbf{H@5} & \textbf{MRR} & \textbf{H@5} & \textbf{MRR} & \textbf{H@5} & \textbf{MRR} & \textbf{H@5}\\
\midrule
\multicolumn{11}{l}{\textbf{\textit{Graph Only (Static Graph)}}} \\
GCN & $67.4{\scriptstyle\pm1.9}$ & $82.3{\scriptstyle\pm1.3}$ & $77.8{\scriptstyle\pm0.6}$ & $89.4{\scriptstyle\pm0.3}$ & $76.4{\scriptstyle\pm0.5}$ & $89.6{\scriptstyle\pm0.5}$ & $48.9{\scriptstyle\pm3.9}$ & $66.6{\scriptstyle\pm3.3}$ & $17.3{\scriptstyle\pm1.5}$ & $23.3{\scriptstyle\pm1.9}$ \\
GAT & $65.4{\scriptstyle\pm1.6}$ & $80.1{\scriptstyle\pm1.0}$ & $78.1{\scriptstyle\pm0.6}$ & $89.3{\scriptstyle\pm0.3}$ & $76.6{\scriptstyle\pm0.7}$ & $90.8{\scriptstyle\pm0.6}$ & $44.7{\scriptstyle\pm0.7}$ & $62.5{\scriptstyle\pm0.7}$ & $16.2{\scriptstyle\pm2.3}$ & $18.0{\scriptstyle\pm2.1}$ \\
TFConv & $68.5{\scriptstyle\pm0.2}$ & $83.2{\scriptstyle\pm0.4}$ & $80.5{\scriptstyle\pm0.2}$ & $90.9{\scriptstyle\pm0.2}$ & $82.4{\scriptstyle\pm0.1}$ & $94.4{\scriptstyle\pm0.0}$ & $55.8{\scriptstyle\pm0.7}$ & $71.5{\scriptstyle\pm0.4}$ & $18.8{\scriptstyle\pm0.7}$ & $25.2{\scriptstyle\pm0.3}$ \\
\addlinespace[0.6ex]
\multicolumn{11}{l}{\textbf{\textit{Two-Stage (Static Graph + Sequence)}}} \\
GCN + S & $68.6{\scriptstyle\pm1.1}$ & $83.4{\scriptstyle\pm0.7}$ & $79.8{\scriptstyle\pm0.7}$ & $90.5{\scriptstyle\pm0.4}$ & $76.8{\scriptstyle\pm1.1}$ & $90.1{\scriptstyle\pm0.6}$ & $50.8{\scriptstyle\pm0.8}$ & $68.1{\scriptstyle\pm0.9}$ & $13.5{\scriptstyle\pm2.7}$ & $18.2{\scriptstyle\pm2.8}$ \\
GAT + S & $67.9{\scriptstyle\pm0.5}$ & $82.3{\scriptstyle\pm0.5}$ & $80.0{\scriptstyle\pm0.7}$ & $90.3{\scriptstyle\pm0.5}$ & $77.2{\scriptstyle\pm0.5}$ & $90.9{\scriptstyle\pm0.4}$ & $50.3{\scriptstyle\pm0.3}$ & $67.2{\scriptstyle\pm0.4}$ & $14.1{\scriptstyle\pm4.6}$ & $17.2{\scriptstyle\pm5.6}$ \\
TFConv + S & $72.7{\scriptstyle\pm0.3}$ & $86.3{\scriptstyle\pm0.1}$ & $82.5{\scriptstyle\pm0.5}$ & $92.0{\scriptstyle\pm0.3}$ & $83.6{\scriptstyle\pm0.1}$ & $95.1{\scriptstyle\pm0.1}$ & $58.4{\scriptstyle\pm0.2}$ & $73.5{\scriptstyle\pm0.1}$ & $17.7{\scriptstyle\pm1.1}$ & $24.2{\scriptstyle\pm1.1}$ \\
\addlinespace[0.6ex]
\multicolumn{11}{l}{\textbf{\textit{\ourmodel}}} \\
\ourmodel-GCN & $\bm{92.2{\scriptstyle\pm0.6}}$ & $\bm{94.8{\scriptstyle\pm0.5}}$ & $\bm{87.9{\scriptstyle\pm0.2}}$ & $\bm{93.3{\scriptstyle\pm0.1}}$ & $\bm{90.4{\scriptstyle\pm0.4}}$ & $\bm{97.6{\scriptstyle\pm0.2}}$ & $\bm{80.2{\scriptstyle\pm0.2}}$ & $\bm{87.1{\scriptstyle\pm0.1}}$ & $52.3{\scriptstyle\pm2.9}$ & $\underline{72.5{\scriptstyle\pm2.2}}$ \\
\ourmodel-GAT & $78.8{\scriptstyle\pm0.4}$ & $89.5{\scriptstyle\pm0.1}$ & $\underline{83.4{\scriptstyle\pm0.1}}$ & $\underline{92.2{\scriptstyle\pm0.1}}$ & $\underline{89.1{\scriptstyle\pm0.0}}$ & $\underline{97.4{\scriptstyle\pm0.0}}$ & $\underline{75.4{\scriptstyle\pm0.3}}$ & $\underline{86.0{\scriptstyle\pm0.1}}$ & $\bm{63.7{\scriptstyle\pm0.8}}$ & $\bm{75.2{\scriptstyle\pm0.2}}$ \\
\ourmodel-TFConv & $78.4{\scriptstyle\pm0.7}$ & $89.3{\scriptstyle\pm0.4}$ & $82.9{\scriptstyle\pm0.1}$ & $92.1{\scriptstyle\pm0.1}$ & $88.6{\scriptstyle\pm0.4}$ & $97.2{\scriptstyle\pm0.1}$ & $75.1{\scriptstyle\pm0.2}$ & $85.7{\scriptstyle\pm0.2}$ & $\underline{54.0{\scriptstyle\pm0.5}}$ & $68.0{\scriptstyle\pm0.7}$ \\
\bottomrule
\end{tabular}
}
\end{table*}

\paragraph{Evaluation Metrics.} We use ranking-based metrics: MRR and Hits@$k$, following prior relationship/link prediction benchmarks~\citep{NEURIPS2024_fda026cf,huang2023temporal,yi2025tgbseqbenchmarkchallengingtemporal}. The main paper reports both MRR and Hits@5, while the full Hits@$k$ results are available in Appendix \ref{app:exp_details}. Each positive relation is ranked against 100 uniformly sampled non-neighbor users. All results are averaged over three independent runs.

\paragraph{Results.}
In \Cref{tab:link_pred_full}, the results support the value of integrating sequential and relational information when modeling user behavior. Two-stage methods' sequence embeddings generally improve over graph-only baselines, showing that users' individual event sequences contain signals that are not captured by the static graph alone. However, training the graph and sequence model separately leaves substantial gains unrealized: \ourmodel variants obtain the best results across all five datasets. We read the Amazon relationship-prediction gains with caution: an edge there is defined by users co-reviewing at least three products while product identities also appear in the input sequences, so the target is partially recoverable from the inputs. To bound this shortcut, a training-free baseline that ranks candidates by the number of observed products they share with the query user reaches $61.5$/$35.4$/$25.1$ MRR on Movies/Electronics/Clothing, where $71.6$/$43.0$/$29.8\%$ of positive pairs share at least one observed product (\cref{tab:amazon_overlap}); \ourmodel exceeds it by $27$--$45$ MRR points, so shared products carry substantial signal but do not reproduce \ourmodel's performance. Brightkite and Gowalla, whose friendship edges are not a function of the check-in tokens, are even cleaner testbeds, and it is there, especially the large Brightkite gain, that we place the most weight.

\subsection{Fraud Detection}\label{sec:exp_fraud}

\paragraph{Experimental Setup.} 
We evaluate fraud detection task on Amazon review data~\citep{McAuley2015image}. Each user's sequence contains product IDs, ratings, and review text. Each review may receive ``helpful'' or ``unhelpful'' votes from other users. 
Following \citet{dou2020enhancing} and \citet{fathonyintegrating}, we use this helpfulness information as a proxy for fraudulent review incidents. Specifically, we label a user with at least 5 reviews as fraudulent if more than $70\%$ of their reviews are marked unhelpful. Static relation edges connect users who co-reviewed at least three products. Dataset statistics are in \Cref{tab:fraud_data_stats}.

\paragraph{Main Compared Methods.}
As in relationship prediction, the main comparison uses models aligned with the static-relation formulation. We evaluate graph-only models, Sentence-BERT as a text/sequence-only comparator, static graph + text embedding models as frozen two-stage comparators, and end-to-end \ourmodel variants.

\begin{table*}[t!]
\centering
\small
\setlength{\tabcolsep}{3pt}
\caption{Fraud detection performance across three Amazon categories (mean$\pm$std over three seeds, \%; higher is better). All methods share one canonical 60/20/20 user split and are reported at the validation-PR-AUC-selected checkpoint, with PR-AUC computed as $\mathrm{auc}(\mathrm{recall},\mathrm{precision})$. We compare static graph, temporal graph, sequence/text-only (Sentence-BERT), frozen two-stage (static graph + text), \ourmodel, and relational deep learning (RelGT) run \emph{both} without text information and with the same pooled Sentence-BERT review embedding fed to the two-stage baseline.}
\resizebox{0.74\textwidth}{!}{%
\begin{tabular}{lcccccc}
\toprule
 & \multicolumn{2}{c}{\textbf{Amazon-Movies}} & \multicolumn{2}{c}{\textbf{Amazon-Electronics}} & \multicolumn{2}{c}{\textbf{Amazon-Clothing}} \\
\cmidrule(lr){2-3}\cmidrule(lr){4-5}\cmidrule(lr){6-7}
\textbf{Method} & \textbf{Max F1} & \textbf{PR AUC} & \textbf{Max F1} & \textbf{PR AUC} & \textbf{Max F1} & \textbf{PR AUC} \\
\midrule
\multicolumn{7}{l}{\textbf{\textit{Static Graph Methods}}} \\
GCN & $43.4\,{\scriptstyle \pm 0.2}$ & $30.9\,{\scriptstyle \pm 0.2}$ & $12.6\,{\scriptstyle \pm 0.2}$ & $8.7\,{\scriptstyle \pm 0.4}$ & $5.3\,{\scriptstyle \pm 0.4}$ & $2.8\,{\scriptstyle \pm 0.3}$ \\
GAT & $43.4\,{\scriptstyle \pm 0.2}$ & $36.2\,{\scriptstyle \pm 0.6}$ & $12.6\,{\scriptstyle \pm 0.1}$ & $7.1\,{\scriptstyle \pm 0.1}$ & $4.9\,{\scriptstyle \pm 0.3}$ & $2.9\,{\scriptstyle \pm 0.4}$ \\
TFConv & $44.1\,{\scriptstyle \pm 0.1}$ & $39.1\,{\scriptstyle \pm 0.5}$ & $12.8\,{\scriptstyle \pm 0.0}$ & $9.4\,{\scriptstyle \pm 0.5}$ & $4.9\,{\scriptstyle \pm 0.2}$ & $2.8\,{\scriptstyle \pm 0.2}$ \\
\addlinespace[0.8ex]
\multicolumn{7}{l}{\textbf{\textit{Temporal Graph Methods}}} \\
TGN & $75.5\,{\scriptstyle \pm 0.6}$ & $83.0\,{\scriptstyle \pm 0.4}$ & $46.8\,{\scriptstyle \pm 0.8}$ & $\underline{47.1\,{\scriptstyle \pm 1.7}}$ & $34.7\,{\scriptstyle \pm 2.6}$ & $33.4\,{\scriptstyle \pm 3.0}$ \\
DyRep & $74.6\,{\scriptstyle \pm 0.9}$ & $82.3\,{\scriptstyle \pm 0.3}$ & $46.4\,{\scriptstyle \pm 0.5}$ & $46.0\,{\scriptstyle \pm 0.9}$ & $34.2\,{\scriptstyle \pm 2.3}$ & $32.9\,{\scriptstyle \pm 3.3}$ \\
\addlinespace[0.8ex]
\multicolumn{7}{l}{\textbf{\textit{Static Graph + Text Embedding (Frozen, Two-Stage)}}} \\ 
GCN & $70.0\,{\scriptstyle \pm 0.5}$ & $74.8\,{\scriptstyle \pm 0.7}$ & $40.5\,{\scriptstyle \pm 0.5}$ & $38.8\,{\scriptstyle \pm 1.2}$ & $34.5\,{\scriptstyle \pm 1.7}$ & $33.1\,{\scriptstyle \pm 1.6}$ \\
GAT & $73.3\,{\scriptstyle \pm 1.2}$ & $78.7\,{\scriptstyle \pm 1.1}$ & $41.8\,{\scriptstyle \pm 0.4}$ & $40.1\,{\scriptstyle \pm 1.1}$ & $34.3\,{\scriptstyle \pm 1.8}$ & $32.6\,{\scriptstyle \pm 2.0}$ \\
TFConv & $75.5\,{\scriptstyle \pm 0.7}$ & $82.9\,{\scriptstyle \pm 0.5}$ & $43.2\,{\scriptstyle \pm 0.5}$ & $42.4\,{\scriptstyle \pm 1.1}$ & $34.9\,{\scriptstyle \pm 2.5}$ & $32.7\,{\scriptstyle \pm 2.2}$ \\
\addlinespace[0.8ex]
\multicolumn{7}{l}{\textbf{\textit{Relational Deep Learning}}} \\
RelGT & $63.2\,{\scriptstyle \pm 1.0}$ & $68.7\,{\scriptstyle \pm 1.1}$ & $26.7\,{\scriptstyle \pm 1.3}$ & $20.3\,{\scriptstyle \pm 0.3}$ & $13.1\,{\scriptstyle \pm 0.8}$ & $6.6\,{\scriptstyle \pm 0.6}$ \\
\addlinespace[0.8ex]
\multicolumn{7}{l}{\textbf{\textit{Relational Deep Learning + Text Embedding}}} \\
RelGT & $77.7\,{\scriptstyle \pm 0.7}$ & $85.0\,{\scriptstyle \pm 0.6}$ & $45.5\,{\scriptstyle \pm 0.5}$ & $45.2\,{\scriptstyle \pm 0.7}$ & $\bm{38.3\,{\scriptstyle \pm 2.1}}$ & $\bm{35.6\,{\scriptstyle \pm 2.3}}$ \\
\addlinespace[0.8ex]
\multicolumn{7}{l}{\textbf{\textit{Sequence Only (Sentence-BERT)}}} \\
SBert & $75.2\,{\scriptstyle \pm 0.8}$ & $82.4\,{\scriptstyle \pm 0.4}$ & $43.5\,{\scriptstyle \pm 0.3}$ & $41.1\,{\scriptstyle \pm 1.1}$ & $33.7\,{\scriptstyle \pm 1.0}$ & $32.1\,{\scriptstyle \pm 1.9}$ \\
\addlinespace[0.8ex]
\multicolumn{7}{l}{\textbf{\textit{\ourmodel}}} \\
\ourmodel-GCN & $\underline{79.6\,{\scriptstyle \pm 0.5}}$ & $\underline{87.2\,{\scriptstyle \pm 0.3}}$ & $\underline{49.0\,{\scriptstyle \pm 0.3}}$ & $\bm{48.7\,{\scriptstyle \pm 0.6}}$ & $36.2\,{\scriptstyle \pm 2.2}$ & $34.6\,{\scriptstyle \pm 2.4}$ \\
\ourmodel-GAT & $79.4\,{\scriptstyle \pm 0.1}$ & $86.5\,{\scriptstyle \pm 0.2}$ & $47.6\,{\scriptstyle \pm 0.5}$ & $46.4\,{\scriptstyle \pm 0.9}$ & $\underline{37.3\,{\scriptstyle \pm 2.2}}$ & $\underline{35.0\,{\scriptstyle \pm 2.3}}$ \\
\ourmodel-TFConv & $\bm{79.9\,{\scriptstyle \pm 1.0}}$ & $\bm{87.6\,{\scriptstyle \pm 0.5}}$ & $\bm{49.5\,{\scriptstyle \pm 0.8}}$ & $\bm{48.7\,{\scriptstyle \pm 1.9}}$ & $36.7\,{\scriptstyle \pm 1.7}$ & $34.7\,{\scriptstyle \pm 2.0}$ \\
\bottomrule
\end{tabular}
}
\label{tab:fraud_three}
\end{table*}

\begin{enumerate}
    \item \textit{\textbf{Graph-Only Models.}}
    GCN, GAT, and TFConv operate on a static co-review graph without additional inputs.

    \item \textit{\textbf{Text Embedding Model.}} 
    We encode each review with Sentence-BERT \cite{reimers-2019-sentence-bert}
    to obtain a sequence of text embeddings per user.
    For classification, we then average the embeddings to form a single user representation, which is passed to a classifier.

    \item \textit{\textbf{Static Graph + Text Embedding.}}
    We incorporate the averaged text embeddings per user into static graph models as additional node features. This strategy is similar to the two-stage modeling in the previous task.

    \item \textit{\textbf{\ourmodel:}} 
    Our approach takes sequence of Sentence-BERT embeddings of the review texts combined with a learnable rating embedding as input, passing them through the sequential module of our architecture. The rating embedding, sequential module, and graph module are trained jointly for fraud detection, allowing the model to integrate textual, sequential, and relational signals in an end-to-end manner.

\end{enumerate}

\paragraph{Evaluation Metrics.} 
For the fraud detection task, we report two classification metrics:
\textbf{Max F1} (the best F1 over decision thresholds, with the threshold selected on the \emph{validation} set and applied unchanged to the test set) and \textbf{PR-AUC} (area under the precision–recall curve).

\paragraph{Results.} 

\Cref{tab:fraud_three} shows that review text is the dominant signal for fraud detection: graph-only models underperform substantially (co-review structure alone is weak), while every text-aware method (Sentence-BERT, the frozen two-stage variants, text-fed RelGT, and \ourmodel) clusters within a few PR-AUC points. \ourmodel attains the best PR-AUC on Movies and Electronics ($87.6$ and $48.7$, ahead of the next method), and on Clothing it is statistically tied with text-fed RelGT ($34.6$--$35.0$ vs.\ $35.6$ PR-AUC, overlapping confidence intervals; a paired per-user bootstrap over identical test users confirms no consistent difference: across-seed gap $+0.7$ PR-AUC / $+0.2$ Max-F1, sign-flipping across seeds, \cref{app:significance}). Two properties of Clothing make this tie \emph{unresolvable} rather than a demonstrated equality: fraud positives are scarce (a $2.4\%$ positive rate, $n\!\approx\!1{,}658$ labeled fraud users, versus $27.7\%$ and $6.5\%$ on Movies and Electronics), so PR-AUC carries a wide seed CI and a sub-point gap cannot be separated from noise; and Clothing reviews are markedly thinner (median $47$ vs.\ $142$ and $86$ tokens per review in our data), leaving less across-review variation for an ordered encoder to exploit over pooling. Because PR-AUC is anchored to prevalence, absolute values are not comparable across categories: Clothing's $\sim$$35$ is roughly $14\times$ its $2.4\%$ base rate. Fraud is throughout a largely \emph{node-local}, text-dominated task (the co-review graph is weak in every category), whereas the regime where relational structure and joint training are decisive is the relationship-prediction task (\cref{sec:exp_friend}).

\paragraph{Relational Deep Learning comparison.}
A natural question is whether Relational Deep Learning (RDL)~\citep{fey_relational_2023}, which also learns end-to-end over a graph induced from the data, already subsumes this setting. RDL is \emph{applicable}: a single user-to-user relation is a (degenerate) relational schema. We cast each Amazon category as a RelBench-style entity graph (users, products, and reviews linked by keys) and predict the same fraud label with the Relational Graph Transformer (RelGT)~\citep{dwivedi2025relgt} in two configurations: with categorical attributes only, and with the \emph{same} pooled Sentence-BERT review embedding fed to our two-stage baseline (\cref{tab:fraud_three}). The comparison is instructive on two fronts. First, RelGT with categorical attributes alone falls far below every text-aware method (e.g.\ $6.6$ PR-AUC on Clothing vs.\ $32.1$ for Sentence-BERT), confirming that the review \emph{text}, not the relational scaffolding, carries the fraud signal. Second, once RelGT is given the same text it becomes competitive: it is statistically tied with \ourmodel on Clothing ($35.6$ vs.\ $34.6$--$35.0$ PR-AUC, overlapping CIs; a comparison underpowered by Clothing's scarce fraud positives, $2.4\%$) and trails only modestly on Movies and Electronics ($85.0$ vs.\ $87.6$; $45.2$ vs.\ $48.7$). We read this as consistent with our thesis rather than counter to it: fraud is a largely \emph{node-local}, text-dominated task, so once the per-user text is available, how it is encoded and routed, an ordered sequence with graph message passing (\ourmodel) versus unordered typed tokens reached by $k$-hop sampling (RDL), is second-order. The decisive advantage on the relationship-prediction task instead comes from jointly training the sequence and graph rather than freezing the encoder (\cref{sec:exp_friend}), where the signal is relational and cold-start rather than node-local.

\subsection{When Does End-to-End Training Help?}\label{sec:mechanism}

The gains above are consistent across the relationship-prediction datasets and shrink on the node-local fraud task, where the co-review graph is weak (\cref{tab:fraud_three}). The two-stage-vs-\ourmodel comparison in \Cref{tab:link_pred_full} shows that end-to-end training improves over the frozen two-stage pipeline precisely where the relation graph carries signal, with the largest gain on Brightkite, whose friendship edges are not recoverable from the check-in tokens. The practical lesson is that jointly training the two structures is worth its cost where the graph carries signal and offers little where the predictive signal is node-local.

\paragraph{Matched freezing control.} The comparison above changes more than encoder updating: the frozen baseline's encoder is pretrained with masked-token prediction, whereas \ourmodel starts from random initialization and is supervised by the task. To isolate encoder updating, we start both pipelines on Brightkite from the same masked-token-pretrained checkpoint and change only whether task gradients update the encoder (\cref{tab:matched_freeze}). Fine-tuning raises MRR from $76.8$ to $87.4$ ($+10.6$). Fine-tuning costs $2.2\times$ the training time per epoch and $1.9\times$ the peak GPU memory of the frozen pipeline (\cref{tab:cost}); we compare per-epoch time because early stopping ends the runs at different epochs, and we time training only, not inference or the reuse of frozen embeddings. Removing relational message passing altogether lowers MRR from $85.4$ to $62.0$ on Brightkite and from $87.9$ to $50.8$ on Gowalla (\cref{tab:ablate_relational}), so the gains also depend on propagating information over the graph. \Cref{app:threats} discusses the scope of these conclusions.

\begin{table}[t]
\centering
\small
\setlength{\tabcolsep}{5pt}
\caption{Matched freezing control on Brightkite (mean$\pm$std over three seeds). The first two rows share the same masked-token-pretrained encoder, architecture, data, and downstream settings (batch size 64) and differ only in whether the encoder is updated. The last row is the main-table \ourmodel-GCN (\cref{tab:link_pred_full}), trained from random initialization with its own selected configuration (batch size 128). The Brightkite no-relation ablation (\cref{tab:ablate_relational}) uses a third, reduced protocol (batch size 16, 30 epochs), hence its lower MRR.}
\label{tab:matched_freeze}
\begin{tabular}{lccc}
\toprule
\textbf{Training setup} & \textbf{MRR} & \textbf{Hits@1} & \textbf{Hits@5} \\
\midrule
MLM-pretrained $\rightarrow$ frozen encoder $\rightarrow$ GNN & $76.8{\scriptstyle\pm0.7}$ & $68.2{\scriptstyle\pm0.8}$ & $87.4{\scriptstyle\pm0.5}$ \\
Same checkpoint $\rightarrow$ encoder fine-tuned with GNN & $87.4{\scriptstyle\pm0.4}$ & $82.9{\scriptstyle\pm0.5}$ & $92.9{\scriptstyle\pm0.2}$ \\
Random init.\ $\rightarrow$ end-to-end (\ourmodel) & $92.2{\scriptstyle\pm0.6}$ & $89.9{\scriptstyle\pm0.7}$ & $94.8{\scriptstyle\pm0.5}$ \\
\bottomrule
\end{tabular}
\end{table}

\begin{table}[t]
\centering
\small
\setlength{\tabcolsep}{5pt}
\caption{Training cost of the matched freezing control on Brightkite (one NVIDIA L40S; mean$\pm$std over three seeds where applicable). The MLM pretraining is shared; embedding generation includes process start-up.}
\label{tab:cost}
\begin{tabular}{lcc}
\toprule
\textbf{Cost} & \textbf{Pretrained + frozen} & \textbf{Pretrained + fine-tuned} \\
\midrule
MLM pretraining (one-time) & $25.0$ min & $25.0$ min \\
Embedding generation (one-time) & $\le 2.6$ min & not required \\
Downstream training per epoch & $29.8\pm0.1$ min & $65.9\pm1.4$ min \\
Peak GPU memory & $21.3\pm0.1$ GiB & $40.9\pm0.2$ GiB \\
\bottomrule
\end{tabular}
\end{table}

\section{Conclusion}\label{sec:concl}

We investigated whether and when it pays to train a sequence encoder and a graph model \emph{end-to-end} rather than in the standard two-stage (frozen-embedding) pipeline, for users described by both an event history and a persistent user-to-user relation. End-to-end training improves clearly over frozen pipelines on relationship prediction, where the relation graph appears to carry much of the signal, whereas the gains shrink where the predictive signal is largely node-local (fraud detection). On Brightkite, a matched control that differs only in whether the pretrained encoder is updated attributes a $10.6$-point MRR gain to end-to-end updating, at $2.2\times$ the training time per epoch and $1.9\times$ the peak memory; freezing may still be preferable when embeddings are reused across tasks or precomputed for serving. We also map when this sequence--graph formulation is preferable to temporal-graph and relational-transformer alternatives. The practical lesson is concrete: for asynchronous per-user histories over a persistent relation, jointly training the two structures is worth it where the relation graph carries signal, a regime we make explicit.

\clearpage   %
\bibliographystyle{unsrtnat}
\bibliography{manual_ref,references}

\appendix
\newpage

\section{Additional Related Works}\label{sec:add_related_works}

\subsection{Sequence Modeling}\label{sec:related_seq}

Transformer-based sequence models have delivered state-of-the-art performance across a wide range of language and vision tasks by learning contextual representations over token sequences~\citep{vaswani_attention_2017,dai-etal-2019-transformer,devlin2019bert,bubeck2023sparksartificialgeneralintelligence}. Given an input sequence, these models map each token to a continuous embedding and use self-attention to capture local and long-range dependencies within the sequence. While highly effective for modeling the internal structure of a single sequence, this paradigm typically treats each sequence in isolation and does not leverage information from \emph{related} sequences.
\subsection{Graph Modeling}\label{sec:related_graph}

Graph models capture relational structures among entities (nodes). In particular, graph neural networks (GNNs) propagate and aggregate information over local neighborhoods to learn node, edge, or graph level embeddings~\citep{kipf2017semi,velickovic_graph_2018,ijcai2021p214}. These methods have demonstrated strong performance across tasks such as node classification\citep{kipf2017semi,10.5555/3294771.3294869}, link prediction~\citep{kipf2016variationalgraphautoencoders,zhang2018linkpredictionbasedgraph}, and recommendation~\citep{Ying_2018,Wang_2019,wu2022graphneuralnetworksrecommender}. Standard GNNs model each node in the graph as a single feature vector and do not account for sequential information within nodes. Bridging graph-based relational modeling with sequence-aware representations is therefore an important step toward richer models that capture both intra-sequence patterns and inter-sequence relations.

\subsection{Temporal Graph Methods}
Temporal graph models such as TGN~\citep{tgn_icml_grl2020}, TGAT~\citep{tgat_iclr20}, DyRep~\citep{trivedi2019dyrep},TNCN~\citep{zhang2025efficient}, and others~\citep{luo2022neighborhood,pmlr-v235-gravina24a,lu2024improving,wang2021tcltransformerbaseddynamicgraph,cong2023do,Liu2024SelfGNNSGB,Tiwari2025HeterogeneousSGF}, attach timestamps to edges to model interactions without aligned time steps. However, a temporal graph requires all events to be described as a relation between two entities. Personal events that involve only a single user cannot be easily represented by a temporal graph. A common workaround is to introduce event nodes (e.g., representing ``login'' or ``payment declined'' as nodes) and connect them to the corresponding entities, but this can create high-degree hubs (e.g., a single ``login'' node connected to many users) and is often unnatural in a graph formulation. 
As argued by \citet{bechler-speicher2025position}, forcing an unsuitable data modality into a graph form can fail to encode important information and may not be meaningful.

\subsection{Spatial-Temporal Graph 
Methods}
Spatial-temporal graph networks were originally designed for time-series prediction problems such as traffic forecasting. Models like STGCN~\citep{yu_spatio-temporal_2018}, STGformer~\citep{wang_stgformer_2024}, MTGCN~\citep{wu_connecting_2020}, and MST-GAT~\citep{DING2023527}, among others~\citep{hafez2021convdysatdeepneuralrepresentation,Bentsen2023ItIAC,Fan2025TowardsMSD}, encode spatial relations (e.g., road networks) and temporal patterns. In recommendation systems, hybrid approaches have also emerged. \citet{shui_sequence-graph_2022} models the relational graph among items while simultaneously capturing user consumption sequences. Similarly, \citet{zhang_dynamic_2023} treats users and items as different node types within a heterogeneous graph, enabling message passing between items (sequential) and users (relational). A common limitation of these methods is the assumption that all time series are synchronized and share identical time steps. By contrast, our setting imposes no such constraint: sequences may be asynchronous, vary in length, and even be non-overlapping. This flexibility better reflects real-world scenarios where user interactions occur at irregular intervals and exhibit diverse temporal patterns across different users.

\subsection{Graph Large Language Models}
An emerging line of research seeks to leverage the generalization capabilities of foundation models for graph learning. \emph{Graph large language models (Graph LLMs)} achieve this by integrating pretrained LLMs with graph-structured data through prompting~\citep{Chen2023ExploringTPA,Perozzi2024LetYGA,Tian2023GraphNPA}, instruction tuning~\citep{Wang2025ExploringGTA,Ye2023LanguageIAA,Guo2024GraphEditLLA,Tan2024GraphorientedITA}, or hybrid architectures~\citep{Ren2024ASOA,Huang2024LargeLMA,Liu2025GraphMLLMHMA}.
The promise of Graph LLMs lies in bridging reasoning over relational data with the flexibility of natural language interfaces. In line with this view, \citet{Zhou2025EachGIA} even conceptualizes that each graph can be seen as a new language for the LLMs to learn. Our study differs in aim: rather than adapting pretrained LLMs to graphs, we hold a standard sequence-encoder-and-GNN pipeline fixed and ask when training it end-to-end helps, treating the architecture as a testbed rather than a contribution.

\subsection{Formulation comparison}\label{app:formulation}
\Cref{tab:formulation_fit} tabulates how each formulation models a user history and a relationship and where it fits best; the trade-offs are discussed in \cref{sec:formulations}.

\begin{table*}[t]
\centering
\small
\setlength{\tabcolsep}{8pt}
\renewcommand{\arraystretch}{1.3}
\caption{How each formulation models a user history and a relationship, and where it fits best. \textbf{E2E}: trained end-to-end on the downstream task. \textbf{Seq.}: the sequence is encoded as a first-class object.}
\label{tab:formulation_fit}
\resizebox{\textwidth}{!}{%
\begin{tabular}{@{}>{\raggedright\arraybackslash}p{0.14\linewidth} >{\raggedright\arraybackslash}p{0.25\linewidth} >{\raggedright\arraybackslash}p{0.23\linewidth} cc >{\raggedright\arraybackslash}p{0.26\linewidth}@{}}
\toprule
\textbf{Formulation} & \textbf{User history} & \textbf{Relationship} & \textbf{E2E} & \textbf{Seq.} & \textbf{Best-fit setting} \\
\midrule
Temporal graph &
\begin{itemize}[nosep]
\item Timestamped edges
\item Per-node memory state
\end{itemize} &
\begin{itemize}[nosep]
\item Dynamic edge
\item E.g., A pays B
\end{itemize} &
\checkmark & $\times$ &
\begin{itemize}[nosep]
\item High-frequency peer-to-peer events
\item Relations form/dissolve over time
\end{itemize} \\
RDL / Rel.\ graph transformer &
\begin{itemize}[nosep]
\item Neighborhood of child rows, e.g., transactions near a user
\end{itemize} &
\begin{itemize}[nosep]
\item Primary/foreign-key link
\item Fixed by the DB schema
\end{itemize} &
\checkmark & $\times$ &
\begin{itemize}[nosep]
\item Enterprise ML on relational DBs
\item Rich, multi-table schemas
\end{itemize} \\
Sequence--graph (ours) &
\begin{itemize}[nosep]
\item Independent, ordered sequence
\item Per-user event tokens
\end{itemize} &
\begin{itemize}[nosep]
\item Separate static graph
\item Gates which sequences interact
\end{itemize} &
\checkmark & \checkmark &
\begin{itemize}[nosep]
\item Dense histories + social structure, e.g., relationship prediction
\item Weaker if node-local (\cref{sec:exp_fraud})
\end{itemize} \\
\bottomrule
\end{tabular}
}
\end{table*}

\section{Algorithms}\label{app:algo}

In this section, we provide the detailed procedural steps for the \ourmodel framework. \Cref{alg:bridge-family} outlines the end-to-end training procedure, illustrating the joint optimization of the sequence model and the graph module.

\begin{algorithm}[h!]
\caption{\ourmodel}
\label{alg:bridge-family}
\begin{algorithmic}[1]
\REQUIRE Sequences $\{S_i\}_{i=1}^N$;  Graph $G=(\mathcal{V},\mathcal{E})$; number of sequence model layers $L_s$; number of graph layers $L_g$
\ENSURE Final user representations $\{h_i\}_{i=1}^N$
\FORALL{$v_i \in \mathcal{V}$}  \STATE $X_i^{(0)} \gets \text{TokenEmbedding}(S_i)$  \ENDFOR
\FORALL{$\ell=1,\dots,L_s$}
  \FORALL{$v_i \in \mathcal{V}$}
    \STATE $X_i^{(\ell)} \gets \mathrm{TransformerLayer}^{(\ell)}\big(X_i^{(\ell-1)}\big)$ 
  \ENDFOR
\ENDFOR
\FORALL{$v_i \in \mathcal{V}$}  
\STATE $Z_i \gets X_i^{(L_s)}$
\STATE $z_i^{(0)} \gets \text{Compress}(Z_i)$
\ENDFOR
\FORALL{$\ell=1,\dots,L_g$}
  \FORALL{$v_i \in \mathcal{V}$}
    \STATE $z_i^{(\ell)} \gets \mathrm{MessagePassing}^{(\ell)} \big(i,\,z_i^{(\ell-1)},\,\{z_j^{(\ell-1)}\}_{j\in\mathcal{N}(i)}\big)$ 
  \ENDFOR
\ENDFOR
\FORALL{$v_i \in \mathcal{V}$} 
\STATE $h_i \gets z_i^{(L_g)} + \text{Compress}(Z_i)$ \COMMENT{residual}
\ENDFOR
\end{algorithmic}
\end{algorithm}

\section{Experimental Details}\label{app:exp_details}
All experiments are conducted on L40 GPUs with 48GB of memory. 
\subsection{Hyperparameters}\label{app:linkpred_hparams}

\paragraph{Relationship prediction}
In \Cref{tab:linkpred_hparams_graph_only_static,tab:linkpred_hparams_two_stage_static_sequence,tab:linkpred_hparams_temporal_graphs,tab:linkpred_hparams_ourmodel}, we report the hyperparameters used for the relationship prediction experiments.
We use the following notation: LR denotes the learning rate, and BS denotes the batch size. For methods that use user-sequence embeddings (Two-Stage and \ourmodel), Bert specifies the configuration of the \emph{BERT-style sequence encoder} used to produce user embeddings from event sequences. We report Bert as $L_{\text{TF}}\times H$, where $L_{\text{TF}}$ is the number of Transformer encoder layers and $H$ is the number of self-attention heads per layer; for example, \texttt{1x2} denotes $L_\text{TF}=1$ and $H=2$. Hidden denotes the embedding dimension of this encoder. The resulting embeddings are used as node features in Two-Stage baselines, or as the sequence module within \ourmodel.

Across static graph methods, we follow the training setup of \citet{fathonyintegrating} and use a shared learning rate and batch size (\Cref{tab:linkpred_hparams_graph_only_static}). 
For temporal graph baselines, we use $\mathrm{bs}=4096$ by default, but reduce the batch size for methods that exceed GPU memory under this setting; the per-method batch sizes are reported in \Cref{tab:linkpred_hparams_temporal_graphs}. The remaining hyperparameters follow existing works: TGN, DyRep, and TNCN follow \citet{fathonyintegrating}, and GraphMixer, NAT, TCL, and TPNet follow the code of \citet{lu2024improving}.

For \ourmodel, we use a default batch size of 128 (reduced when required by memory constraints) and choose the BERT-style sequence encoder configuration to match the user-embedding encoder used by the Two-Stage static baselines.

\begin{table}[h]
\caption{\textbf{Relationship prediction hyperparameters (graph-only static baselines).} We report the learning rate (LR) and batch size (BS) used for each static graph encoder.}
\label{tab:linkpred_hparams_graph_only_static}
\centering
\small
\setlength{\tabcolsep}{3pt}
\begin{tabular}{lcc}
\toprule
Method & LR & BS \\
\midrule
GAT & 0.001 & 4096 \\
GCN & 0.001 & 4096 \\
TFConv & 0.001 & 4096 \\
\bottomrule
\end{tabular}
\end{table}

\begin{table}[h]
\caption{\textbf{Relationship prediction hyperparameters (two-stage static + sequence baselines).}
We report the learning rate (LR), batch size (BS), and sequence-encoder configuration:
``Bert'' denotes \# Transformer layers $\times$ \#attention heads, and ``Hidden'' is the Transformer hidden size.}
\label{tab:linkpred_hparams_two_stage_static_sequence}
\centering
\small
\setlength{\tabcolsep}{3pt}
\renewcommand{\arraystretch}{0.95}
\begin{tabular}{lcccc}
\toprule
Method & LR & BS & Bert & Hidden \\
\midrule
GAT + S & 0.001 & 4096 & 1x2 & 128 \\
GCN + S & 0.001 & 4096 & 1x2 & 128 \\
TFConv + S & 0.001 & 4096 & 1x2 & 128 \\
\bottomrule
\end{tabular}
\end{table}

\begin{table}[tbh]
\caption{\textbf{Relationship prediction hyperparameters (temporal graph baselines).}
We report the learning rate (LR) and batch size (BS) used for each temporal graph method.}
\label{tab:linkpred_hparams_temporal_graphs}
\centering
\small
\setlength{\tabcolsep}{3pt}
\renewcommand{\arraystretch}{0.95}
\begin{tabular}{lcc}
\toprule
Method & LR & BS \\
\midrule
DyRep & 0.001 & 4096 \\
TGN & 0.001 & 4096 \\
TCL & 0.001 & 1024 \\
NAT & 0.001 & 1024 \\
GraphMixer & 0.001 & 1024 \\
TPNet & 0.001 & 1024 \\
\bottomrule
\end{tabular}
\end{table}

\begin{table}[tbh]
\caption{\textbf{Relationship prediction hyperparameters (\ourmodel).}
For each dataset and graph backbone, we report the learning rate (LR), batch size (BS), and sequence-encoder configuration.
``Bert'' denotes \#Transformer layers $\times$ \#attention heads, and ``Hidden'' is the Transformer hidden size.
``--'' indicates the configuration is not applicable or not run for that dataset/backbone.}
\label{tab:linkpred_hparams_ourmodel}
\centering
\small
\setlength{\tabcolsep}{3pt}
\renewcommand{\arraystretch}{0.95}
\begin{tabular}{llcccc}
\toprule
Method & Dataset & LR & BS & Bert & Hidden \\
\midrule
\multirow{3}{*}{\ourmodel-GCN} & Brightkite & 0.0001 & 128 & 1x2 & 128 \\
& Gowalla & 0.0001 & 32 & 1x2 & 128 \\
 & Amazon's & 0.0001 & 32 & 1x2 & 128 \\
\cmidrule(lr){1-6}
\multirow{3}{*}{\ourmodel-GAT} & Brightkite & 0.0001 & 128 & 1x2 & 128 \\
& Gowalla & 0.0001 & 32 & 1x2 & 128 \\
 & Amazon's & 0.0001 & 32 & 1x2 & 128 \\
 \cmidrule(lr){1-6}
\multirow{3}{*}{\ourmodel-TFConv} & Brightkite & 0.0001 & 128 & 1x2 & 128 \\
& Gowalla & 0.0001 & 32 & 1x2 & 128 \\
 & Amazon's & 0.0001 & 32 & 1x2 & 128 \\
\bottomrule
\end{tabular}

\end{table}

\paragraph{Fraud detection.}
We summarize the hyperparameters for fraud detection in
\Cref{tab:fraud_hparams_static_graph_methods,tab:fraud_hparams_two-stage_temporal_graph_methods,tab:fraud_hparams_bridge_methods_advanced}.
Across method classes, we tune LR and BS, and, when applicable, attention heads and sequence-encoder capacity parameters, select the configuration with the best validation performance, and report test results averaged over multiple random seeds.

\paragraph{Static graph methods.}
\Cref{tab:fraud_hparams_static_graph_methods} lists hyperparameters for graph-only and two-stage graph baselines operating on the interaction graph, with or without a text encoder.
Here, Heads denotes the number of attention heads for attention-based GNNs (not applicable to GCN).

\paragraph{Two-Stage temporal graph methods.}
\Cref{tab:fraud_hparams_two-stage_temporal_graph_methods} reports the selected hyperparameters for Two-Stage temporal graph baselines.
These methods use review-text embeddings as precomputed node features; Heads denotes the number of attention heads for models that use attention.

\paragraph{\ourmodel methods.}
\Cref{tab:fraud_hparams_bridge_methods_advanced} reports the per-dataset configurations for end-to-end \ourmodel variants. In addition to LR BS, and Heads, we report the classifier depth (Clf-Layers) and the sequence Transformer capacity as separate columns: {TF-Layers}, {TF-Heads}, {TF-FFN}, and {TF-Dropout}.
All tabulated settings are extracted from the same run artifacts used to produce the main fraud detection results.

\begin{table}[h]
\caption{\textbf{Fraud detection hyperparameters (static graph baselines).}
We report the learning rate (LR), batch size (BS), and number of attention heads (Heads) used for each static graph method.
``--'' indicates the method is non-attentional and does not use heads.}
\label{tab:fraud_hparams_static_graph_methods}
\centering
\small
\setlength{\tabcolsep}{3pt}
\renewcommand{\arraystretch}{0.95}
\begin{tabular}{lccc}
\toprule
Method & LR & BS & Heads \\
\midrule
GAT & 0.001 & 4096 & 2 \\
GCN & 0.001 & 4096 & -- \\
TFConv & 0.001 & 4096 & 2 \\
\bottomrule
\end{tabular}
\end{table}

\begin{table}[h]
\caption{\textbf{Fraud detection hyperparameters (temporal graph baselines).} We report the learning rate (LR), batch size (BS), and attention heads (Heads) used in the temporal encoder for each method.}

\label{tab:fraud_hparams_two-stage_temporal_graph_methods}
\centering
\small
\setlength{\tabcolsep}{3pt}
\renewcommand{\arraystretch}{0.95}
\begin{tabular}{lccc}
\toprule
Method & LR & BS & Heads \\
\midrule
DyRep & 0.001 & 32768 & 2 \\
TGN & 0.001 & 32768 & 2 \\
TCL & 0.001 & 32768 & 2 \\
NAT & 0.001 & 32768 & 1 \\
GraphMixer & 0.001 & 32768 & 1 \\
TPNet & 0.001 & 32768 & 1 \\
\bottomrule
\end{tabular}
\end{table}

\begin{table*}[h]
\caption{\textbf{Fraud detection hyperparameters (\ourmodel variants).}
For each dataset and Bridge backbone, we list the optimizer learning rate (LR) and batch size (BS), the number of graph attention heads (Heads; ``--'' for non-attentional GCN), the number of classifier MLP layers (Clf-Layers), and Transformer sequence-encoder settings (TF-Layers/Heads, FFN hidden size TF-FFN, and dropout TF-Dropout).}
\label{tab:fraud_hparams_bridge_methods_advanced}
\centering
\scriptsize
\setlength{\tabcolsep}{2pt}
\renewcommand{\arraystretch}{0.95}
\begin{tabular}{@{}llcccccccc@{}}
\toprule
Method & Dataset & LR & BS & Heads & Clf. & TF-L & TF-H & FFN & Drop. \\
\midrule
\multirow{3}{*}{GAT+GAT} & Movies & 0.0003 & 512 & 1 & 2 & 2 & 4 & 512 & 0.2 \\
 & Electronics & 0.0003 & 512 & 1 & 2 & 2 & 4 & 256 & 0.2 \\
 & Clothing & 0.0003 & 512 & 1 & 2 & 2 & 2 & 256 & 0.2 \\
\cmidrule(lr){1-10}
\multirow{3}{*}{GCN+GCN} & Movies & 0.0003 & 512 & -- & 2 & 2 & 4 & 512 & 0.2 \\
 & Electronics & 0.0003 & 512 & -- & 2 & 2 & 4 & 256 & 0.2 \\
 & Clothing & 0.0003 & 512 & -- & 2 & 2 & 2 & 256 & 0.15 \\
\cmidrule(lr){1-10}
\multirow{3}{*}{TFConv+TFConv} & Movies & 0.0003 & 512 & 1 & 2 & 2 & 4 & 512 & 0.2 \\
 & Electronics & 0.0003 & 512 & 1 & 2 & 2 & 4 & 256 & 0.2 \\
 & Clothing & 0.0003 & 512 & 1 & 2 & 2 & 2 & 256 & 0.15 \\
\bottomrule
\end{tabular}

\end{table*}

\subsection{Full Relationship Prediction Results}\label{app:linkpred_full}

\paragraph{Metrics and protocol.}
We evaluate relationship prediction using standard ranking metrics, Mean Reciprocal Rank (MRR) and Hits@$k$ for $k\in\{1,3,5,10\}$, following established benchmarks~\citep{NEURIPS2024_fda026cf,huang2023temporal,yi2025tgbseqbenchmarkchallengingtemporal}.
For each query, we rank the ground-truth target among 101 candidates (one positive and 100 negatives), where negatives are sampled uniformly according to the benchmark protocol and restricted to non-edges for the query.
Positive relations are split into training, validation, and test sets; only training relations enter the message-passing graph, validation and test relations are used only as prediction targets, and no relation of any split is used for sequence pretraining, which sees only users' event sequences.
We report the mean and standard deviation over three independent runs.

\paragraph{Full results.}
\Cref{tab:link_pred_app} reports the complete set of ranking metrics for all methods and datasets.
In the main paper, we present MRR and Hits@$5$ for brevity; the additional Hits@$k$ values reported here provide a more detailed view of performance across operating points and do not change the qualitative conclusions.

\begin{table*}[h]
\centering
\scriptsize
\caption{\textbf{Relationship prediction ranking performance on Brightkite and Gowalla (higher is better).}
We report mean$\pm$std over three seeds.
\textbf{Hits@k} is the fraction of test queries for which at least one ground-truth relation appears in the top-$k$ predictions.
Best results are \textbf{bolded} and runner-ups are \underline{underlined}.}
\label{tab:link_pred_app}
\setlength{\tabcolsep}{4pt}
\resizebox{\textwidth}{!}{%
\begin{tabular}{@{}l*{10}{@{\hspace{0.5pt}}c}@{}}
\toprule
& \multicolumn{5}{c}{\textbf{Brightkite}} & \multicolumn{5}{c}{\textbf{Gowalla}}\\
\cmidrule(lr){2-6}\cmidrule(lr){7-11}
\textbf{Method} & \textbf{MRR} & \textbf{H@1} & \textbf{H@3} & \textbf{H@5} & \textbf{H@10} & \textbf{MRR} & \textbf{H@1} & \textbf{H@3} & \textbf{H@5} & \textbf{H@10}\\
\midrule
\multicolumn{11}{l}{\textbf{\textit{Graph Only (Static)}}} \\
GCN & $67.4{\scriptstyle\pm1.9}$ & $55.6{\scriptstyle\pm2.4}$ & $75.5{\scriptstyle\pm1.8}$ & $82.3{\scriptstyle\pm1.3}$ & $89.2{\scriptstyle\pm0.7}$  & $77.8{\scriptstyle\pm0.6}$ & $68.6{\scriptstyle\pm0.8}$ & $84.9{\scriptstyle\pm0.4}$ & $89.4{\scriptstyle\pm0.3}$ & $93.5{\scriptstyle\pm0.2}$ \\
GAT & $65.4{\scriptstyle\pm1.6}$ & $53.6{\scriptstyle\pm2.0}$ & $73.0{\scriptstyle\pm1.4}$ & $80.1{\scriptstyle\pm1.0}$ & $87.8{\scriptstyle\pm0.6}$  & $78.1{\scriptstyle\pm0.6}$ & $69.2{\scriptstyle\pm0.9}$ & $84.9{\scriptstyle\pm0.4}$ & $89.3{\scriptstyle\pm0.3}$ & $93.3{\scriptstyle\pm0.3}$ \\
TFConv & $68.5{\scriptstyle\pm0.2}$ & $56.9{\scriptstyle\pm0.2}$ & $76.6{\scriptstyle\pm0.4}$ & $83.2{\scriptstyle\pm0.4}$ & $89.8{\scriptstyle\pm0.3}$  & $80.5{\scriptstyle\pm0.2}$ & $72.3{\scriptstyle\pm0.2}$ & $87.0{\scriptstyle\pm0.2}$ & $90.9{\scriptstyle\pm0.2}$ & $94.3{\scriptstyle\pm0.1}$ \\
\addlinespace[0.6ex]
\multicolumn{11}{l}{\textbf{\textit{Two-Stage (Static + Sequence)}}} \\
GCN + S & $68.6{\scriptstyle\pm1.1}$ & $56.9{\scriptstyle\pm1.4}$ & $76.7{\scriptstyle\pm1.0}$ & $83.4{\scriptstyle\pm0.7}$ & $90.0{\scriptstyle\pm0.3}$  & $79.8{\scriptstyle\pm0.7}$ & $71.3{\scriptstyle\pm0.9}$ & $86.5{\scriptstyle\pm0.5}$ & $90.5{\scriptstyle\pm0.4}$ & $94.1{\scriptstyle\pm0.4}$ \\
GAT + S & $67.9{\scriptstyle\pm0.5}$ & $56.4{\scriptstyle\pm0.6}$ & $75.7{\scriptstyle\pm0.6}$ & $82.3{\scriptstyle\pm0.5}$ & $89.4{\scriptstyle\pm0.2}$  & $80.0{\scriptstyle\pm0.7}$ & $71.7{\scriptstyle\pm0.9}$ & $86.3{\scriptstyle\pm0.6}$ & $90.3{\scriptstyle\pm0.5}$ & $93.9{\scriptstyle\pm0.5}$ \\
TFConv + S & $72.7{\scriptstyle\pm0.3}$ & $61.9{\scriptstyle\pm0.4}$ & $80.5{\scriptstyle\pm0.3}$ & $86.3{\scriptstyle\pm0.1}$ & $91.7{\scriptstyle\pm0.1}$  & $82.5{\scriptstyle\pm0.5}$ & $74.8{\scriptstyle\pm0.7}$ & $88.6{\scriptstyle\pm0.4}$ & $92.0{\scriptstyle\pm0.3}$ & $95.0{\scriptstyle\pm0.2}$ \\
\addlinespace[0.6ex]
\multicolumn{11}{l}{\textbf{\textit{Temporal Graphs}}} \\
DyRep & $26.4{\scriptstyle\pm3.7}$ & $14.0{\scriptstyle\pm3.1}$ & $28.2{\scriptstyle\pm4.8}$ & $37.5{\scriptstyle\pm5.5}$ & $53.4{\scriptstyle\pm5.5}$  & $27.8{\scriptstyle\pm2.1}$ & $16.6{\scriptstyle\pm1.3}$ & $29.6{\scriptstyle\pm2.5}$ & $37.8{\scriptstyle\pm3.3}$ & $51.2{\scriptstyle\pm4.6}$ \\
TGN & $38.7{\scriptstyle\pm1.6}$ & $24.8{\scriptstyle\pm1.8}$ & $43.8{\scriptstyle\pm1.7}$ & $53.9{\scriptstyle\pm1.3}$ & $68.1{\scriptstyle\pm1.1}$  & $44.8{\scriptstyle\pm2.7}$ & $31.8{\scriptstyle\pm2.6}$ & $50.2{\scriptstyle\pm3.4}$ & $59.3{\scriptstyle\pm3.4}$ & $71.9{\scriptstyle\pm2.7}$ \\
TCL & $20.4{\scriptstyle\pm1.0}$ & $9.9{\scriptstyle\pm0.7}$ & $21.2{\scriptstyle\pm1.2}$ & $29.1{\scriptstyle\pm1.5}$ & $43.0{\scriptstyle\pm2.0}$  & $20.5{\scriptstyle\pm1.7}$ & $11.1{\scriptstyle\pm1.3}$ & $20.4{\scriptstyle\pm2.1}$ & $27.5{\scriptstyle\pm2.5}$ & $39.9{\scriptstyle\pm3.2}$ \\
NAT & $68.2{\scriptstyle\pm7.8}$ & $60.3{\scriptstyle\pm9.7}$ & $71.9{\scriptstyle\pm7.3}$ & $77.0{\scriptstyle\pm5.9}$ & $83.9{\scriptstyle\pm3.7}$  & $59.1{\scriptstyle\pm13.0}$ & $50.6{\scriptstyle\pm14.7}$ & $61.8{\scriptstyle\pm13.5}$ & $68.0{\scriptstyle\pm11.8}$ & $77.0{\scriptstyle\pm8.5}$ \\
GraphMixer & $20.6{\scriptstyle\pm0.5}$ & $8.9{\scriptstyle\pm0.6}$ & $21.6{\scriptstyle\pm0.6}$ & $31.0{\scriptstyle\pm0.4}$ & $46.5{\scriptstyle\pm0.5}$  & $24.9{\scriptstyle\pm0.5}$ & $13.3{\scriptstyle\pm0.4}$ & $26.5{\scriptstyle\pm0.6}$ & $35.5{\scriptstyle\pm0.7}$ & $50.1{\scriptstyle\pm0.7}$ \\
TPNet & $45.0{\scriptstyle\pm0.6}$ & $35.7{\scriptstyle\pm0.8}$ & $47.0{\scriptstyle\pm0.5}$ & $53.6{\scriptstyle\pm0.4}$ & $64.3{\scriptstyle\pm0.4}$  & $51.4{\scriptstyle\pm1.0}$ & $40.8{\scriptstyle\pm0.8}$ & $54.8{\scriptstyle\pm1.2}$ & $62.9{\scriptstyle\pm1.3}$ & $74.4{\scriptstyle\pm1.5}$ \\
TNCN & $53.8{\scriptstyle\pm0.5}$ & $42.9{\scriptstyle\pm0.4}$ & $58.2{\scriptstyle\pm0.6}$ & $65.9{\scriptstyle\pm1.0}$ & $76.0{\scriptstyle\pm1.1}$  & $51.8{\scriptstyle\pm0.7}$ & $40.7{\scriptstyle\pm0.6}$ & $55.9{\scriptstyle\pm0.9}$ & $64.0{\scriptstyle\pm0.8}$ & $75.4{\scriptstyle\pm0.8}$ \\
\addlinespace[0.6ex]
\multicolumn{11}{l}{\textbf{\textit{\ourmodel}}} \\
\ourmodel-GCN & $\underline{92.2{\scriptstyle\pm0.6}}$ & $\underline{89.9{\scriptstyle\pm0.7}}$ & $\underline{93.6{\scriptstyle\pm0.6}}$ & $\underline{94.8{\scriptstyle\pm0.5}}$ & $\underline{96.3{\scriptstyle\pm0.4}}$  & $\bm{87.9{\scriptstyle\pm0.2}}$ & $\bm{83.4{\scriptstyle\pm0.3}}$ & $\bm{91.1{\scriptstyle\pm0.1}}$ & $\bm{93.3{\scriptstyle\pm0.1}}$ & $\bm{95.5{\scriptstyle\pm0.1}}$ \\
\ourmodel-GAT & $78.8{\scriptstyle\pm0.4}$ & $70.2{\scriptstyle\pm0.5}$ & $85.3{\scriptstyle\pm0.3}$ & $89.5{\scriptstyle\pm0.1}$ & $93.4{\scriptstyle\pm0.1}$  & $\underline{83.4{\scriptstyle\pm0.1}}$ & $\underline{76.3{\scriptstyle\pm0.2}}$ & $\underline{89.0{\scriptstyle\pm0.0}}$ & $\underline{92.2{\scriptstyle\pm0.1}}$ & $\underline{95.2{\scriptstyle\pm0.0}}$ \\
\ourmodel-TFConv & $78.4{\scriptstyle\pm0.7}$ & $69.6{\scriptstyle\pm1.0}$ & $84.9{\scriptstyle\pm0.7}$ & $89.3{\scriptstyle\pm0.4}$ & $93.4{\scriptstyle\pm0.3}$  & $82.9{\scriptstyle\pm0.1}$ & $75.5{\scriptstyle\pm0.1}$ & $88.7{\scriptstyle\pm0.1}$ & $92.1{\scriptstyle\pm0.1}$ & $95.1{\scriptstyle\pm0.1}$ \\
\bottomrule
\end{tabular}
}
\end{table*}

\begin{table*}[h]
\centering
\scriptsize
\caption{\textbf{Relationship prediction ranking performance on Amazon-Clothing and Amazon-Electronics (higher is better).}
Mean$\pm$std over three seeds; best bolded, runner-up underlined.}
\setlength{\tabcolsep}{4pt}
\resizebox{\textwidth}{!}{%
\begin{tabular}{@{}l*{10}{@{\hspace{0.5pt}}c}@{}}
\toprule
& \multicolumn{5}{c}{\textbf{Clothing}} & \multicolumn{5}{c}{\textbf{Electronics}}\\
\cmidrule(lr){2-6}\cmidrule(lr){7-11}
\textbf{Method} & \textbf{MRR} & \textbf{H@1} & \textbf{H@3} & \textbf{H@5} & \textbf{H@10} & \textbf{MRR} & \textbf{H@1} & \textbf{H@3} & \textbf{H@5} & \textbf{H@10}\\
\midrule
\multicolumn{11}{l}{\textbf{\textit{Graph Only (Static)}}} \\
GCN & $17.3{\scriptstyle\pm1.6}$ & $9.1{\scriptstyle\pm1.7}$ & $17.4{\scriptstyle\pm2.0}$ & $23.4{\scriptstyle\pm1.9}$ & $34.1{\scriptstyle\pm0.7}$  & $48.9{\scriptstyle\pm3.9}$ & $34.9{\scriptstyle\pm4.4}$ & $56.9{\scriptstyle\pm4.2}$ & $66.6{\scriptstyle\pm3.3}$ & $76.5{\scriptstyle\pm2.2}$ \\
GAT & $15.9{\scriptstyle\pm1.9}$ & $9.9{\scriptstyle\pm2.0}$ & $14.0{\scriptstyle\pm2.5}$ & $17.9{\scriptstyle\pm2.1}$ & $27.5{\scriptstyle\pm1.5}$  & $44.7{\scriptstyle\pm0.7}$ & $30.5{\scriptstyle\pm0.8}$ & $51.8{\scriptstyle\pm0.8}$ & $62.5{\scriptstyle\pm0.7}$ & $74.6{\scriptstyle\pm0.4}$ \\
TFConv & $18.8{\scriptstyle\pm0.5}$ & $10.4{\scriptstyle\pm0.9}$ & $19.1{\scriptstyle\pm0.2}$ & $25.0{\scriptstyle\pm0.4}$ & $35.6{\scriptstyle\pm0.2}$  & $55.8{\scriptstyle\pm0.7}$ & $43.0{\scriptstyle\pm1.0}$ & $63.8{\scriptstyle\pm0.6}$ & $71.5{\scriptstyle\pm0.4}$ & $79.5{\scriptstyle\pm0.1}$ \\
\addlinespace[0.6ex]
\multicolumn{11}{l}{\textbf{\textit{Two-Stage (Static + Sequence)}}} \\
GCN + S & $11.3{\scriptstyle\pm1.5}$ & $3.1{\scriptstyle\pm1.1}$ & $9.7{\scriptstyle\pm2.9}$ & $15.6{\scriptstyle\pm2.8}$ & $30.1{\scriptstyle\pm1.5}$  & $50.8{\scriptstyle\pm0.8}$ & $37.1{\scriptstyle\pm0.9}$ & $58.8{\scriptstyle\pm1.1}$ & $68.1{\scriptstyle\pm0.9}$ & $77.2{\scriptstyle\pm0.5}$ \\
GAT + S & $11.7{\scriptstyle\pm3.5}$ & $5.6{\scriptstyle\pm3.5}$ & $9.3{\scriptstyle\pm4.5}$ & $12.7{\scriptstyle\pm4.5}$ & $23.3{\scriptstyle\pm2.9}$  & $50.3{\scriptstyle\pm0.3}$ & $36.6{\scriptstyle\pm0.3}$ & $58.1{\scriptstyle\pm0.9}$ & $67.2{\scriptstyle\pm0.4}$ & $76.7{\scriptstyle\pm0.3}$ \\
TFConv + S & $19.8{\scriptstyle\pm2.0}$ & $10.3{\scriptstyle\pm1.5}$ & $20.2{\scriptstyle\pm2.9}$ & $27.5{\scriptstyle\pm3.4}$ & $39.6{\scriptstyle\pm4.3}$  & $58.4{\scriptstyle\pm0.2}$ & $46.2{\scriptstyle\pm0.3}$ & $66.2{\scriptstyle\pm0.1}$ & $73.5{\scriptstyle\pm0.1}$ & $80.8{\scriptstyle\pm0.0}$ \\
\addlinespace[0.6ex]
\multicolumn{11}{l}{\textbf{\textit{Temporal Graphs}}} \\
DyRep & $13.3{\scriptstyle\pm0.8}$ & $3.6{\scriptstyle\pm0.6}$ & $11.2{\scriptstyle\pm1.0}$ & $18.9{\scriptstyle\pm1.3}$ & $34.3{\scriptstyle\pm1.5}$  & $17.0{\scriptstyle\pm1.9}$ & $7.0{\scriptstyle\pm1.8}$ & $16.8{\scriptstyle\pm2.4}$ & $23.7{\scriptstyle\pm2.1}$ & $37.0{\scriptstyle\pm2.1}$ \\
TGN & $14.8{\scriptstyle\pm1.4}$ & $4.9{\scriptstyle\pm0.8}$ & $13.4{\scriptstyle\pm2.1}$ & $20.8{\scriptstyle\pm3.0}$ & $36.9{\scriptstyle\pm3.4}$  & $33.8{\scriptstyle\pm1.5}$ & $21.1{\scriptstyle\pm1.3}$ & $37.6{\scriptstyle\pm1.8}$ & $46.5{\scriptstyle\pm2.1}$ & $60.0{\scriptstyle\pm1.8}$ \\
TCL & $14.2{\scriptstyle\pm0.5}$ & $5.3{\scriptstyle\pm0.2}$ & $13.6{\scriptstyle\pm1.1}$ & $20.5{\scriptstyle\pm1.3}$ & $32.9{\scriptstyle\pm1.5}$  & $34.9{\scriptstyle\pm0.2}$ & $22.4{\scriptstyle\pm0.3}$ & $39.2{\scriptstyle\pm0.1}$ & $47.3{\scriptstyle\pm0.1}$ & $59.6{\scriptstyle\pm0.2}$ \\
NAT & $30.6{\scriptstyle\pm7.9}$ & $23.3{\scriptstyle\pm9.1}$ & $32.2{\scriptstyle\pm8.3}$ & $36.8{\scriptstyle\pm7.2}$ & $43.5{\scriptstyle\pm4.1}$  & $62.8{\scriptstyle\pm8.0}$ & $55.7{\scriptstyle\pm9.7}$ & $65.6{\scriptstyle\pm8.0}$ & $70.4{\scriptstyle\pm6.4}$ & $77.0{\scriptstyle\pm3.7}$ \\
GraphMixer & $12.9{\scriptstyle\pm0.3}$ & $4.6{\scriptstyle\pm0.3}$ & $11.7{\scriptstyle\pm0.3}$ & $17.7{\scriptstyle\pm0.4}$ & $30.1{\scriptstyle\pm0.8}$  & $23.1{\scriptstyle\pm0.5}$ & $10.4{\scriptstyle\pm0.5}$ & $24.3{\scriptstyle\pm0.5}$ & $34.3{\scriptstyle\pm0.5}$ & $51.8{\scriptstyle\pm0.5}$ \\
TPNet & $11.4{\scriptstyle\pm0.6}$ & $3.5{\scriptstyle\pm0.3}$ & $9.1{\scriptstyle\pm0.8}$ & $14.8{\scriptstyle\pm1.1}$ & $27.6{\scriptstyle\pm1.3}$  & $22.9{\scriptstyle\pm0.2}$ & $9.9{\scriptstyle\pm0.3}$ & $24.2{\scriptstyle\pm0.3}$ & $34.7{\scriptstyle\pm0.3}$ & $52.9{\scriptstyle\pm0.5}$ \\
TNCN & $31.7{\scriptstyle\pm1.8}$ & $22.5{\scriptstyle\pm1.7}$ & $33.9{\scriptstyle\pm2.5}$ & $39.6{\scriptstyle\pm2.1}$ & $48.1{\scriptstyle\pm1.6}$  & $45.4{\scriptstyle\pm1.1}$ & $35.7{\scriptstyle\pm1.2}$ & $49.8{\scriptstyle\pm1.1}$ & $55.8{\scriptstyle\pm0.9}$ & $63.0{\scriptstyle\pm1.2}$ \\
\addlinespace[0.6ex]
\multicolumn{11}{l}{\textbf{\textit{\ourmodel}}} \\
\ourmodel-GCN & $52.3{\scriptstyle\pm2.9}$ & $37.1{\scriptstyle\pm4.4}$ & $62.5{\scriptstyle\pm2.7}$ & $72.5{\scriptstyle\pm2.2}$ & $\bm{80.6{\scriptstyle\pm1.7}}$  & $\bm{80.2{\scriptstyle\pm0.2}}$ & $\bm{74.5{\scriptstyle\pm0.4}}$ & $\bm{84.3{\scriptstyle\pm0.1}}$ & $\bm{87.1{\scriptstyle\pm0.1}}$ & $\bm{89.9{\scriptstyle\pm0.2}}$ \\
\ourmodel-GAT & $\underline{63.7{\scriptstyle\pm0.8}}$ & $\underline{54.4{\scriptstyle\pm1.1}}$ & $\underline{70.7{\scriptstyle\pm0.4}}$ & $\bm{75.2{\scriptstyle\pm0.2}}$ & $78.9{\scriptstyle\pm0.6}$  & $75.4{\scriptstyle\pm0.3}$ & $66.9{\scriptstyle\pm0.4}$ & $\underline{82.2{\scriptstyle\pm0.1}}$ & $\underline{86.0{\scriptstyle\pm0.1}}$ & $89.0{\scriptstyle\pm0.1}$ \\
\ourmodel-TFConv & $54.0{\scriptstyle\pm0.5}$ & $42.0{\scriptstyle\pm0.8}$ & $59.7{\scriptstyle\pm0.7}$ & $68.0{\scriptstyle\pm0.7}$ & $\underline{79.2{\scriptstyle\pm1.0}}$  & $75.1{\scriptstyle\pm0.2}$ & $66.4{\scriptstyle\pm0.2}$ & $81.7{\scriptstyle\pm0.3}$ & $85.7{\scriptstyle\pm0.2}$ & $\underline{89.3{\scriptstyle\pm0.0}}$ \\
\bottomrule
\end{tabular}
}
\end{table*}

\begin{table*}[h]
\centering
\small
\caption{\textbf{Relationship prediction ranking performance on Amazon-Movies (higher is better).}
Mean$\pm$std over three seeds; best bolded, runner-up underlined.}
\begin{tabular}{lccccc}
\toprule
& \multicolumn{5}{c}{\textbf{Amazon-Movies}} \\
\cmidrule(lr){2-6}
\textbf{Method} & \textbf{MRR} & \textbf{Hits@1} & \textbf{Hits@3} & \textbf{Hits@5} & \textbf{Hits@10}\\
\midrule
\multicolumn{6}{l}{\textbf{\textit{Graph Only (Static)}}} \\
GCN & $76.4{\scriptstyle\pm0.5}$ & $65.8{\scriptstyle\pm0.6}$ & $84.5{\scriptstyle\pm0.7}$ & $89.6{\scriptstyle\pm0.5}$ & $94.6{\scriptstyle\pm0.2}$ \\
GAT & $76.6{\scriptstyle\pm0.7}$ & $65.5{\scriptstyle\pm0.8}$ & $85.5{\scriptstyle\pm0.8}$ & $90.8{\scriptstyle\pm0.6}$ & $95.3{\scriptstyle\pm0.3}$ \\
TFConv & $82.4{\scriptstyle\pm0.1}$ & $73.0{\scriptstyle\pm0.2}$ & $90.5{\scriptstyle\pm0.0}$ & $94.4{\scriptstyle\pm0.0}$ & $97.4{\scriptstyle\pm0.0}$ \\
\addlinespace[0.6ex]
\multicolumn{6}{l}{\textbf{\textit{Two-Stage (Static + Sequence)}}} \\
GCN + S & $76.8{\scriptstyle\pm1.1}$ & $66.3{\scriptstyle\pm1.4}$ & $84.9{\scriptstyle\pm0.9}$ & $90.1{\scriptstyle\pm0.6}$ & $94.9{\scriptstyle\pm0.3}$ \\
GAT + S & $77.2{\scriptstyle\pm0.5}$ & $66.6{\scriptstyle\pm0.7}$ & $85.7{\scriptstyle\pm0.5}$ & $90.9{\scriptstyle\pm0.4}$ & $95.4{\scriptstyle\pm0.3}$ \\
TFConv + S & $83.6{\scriptstyle\pm0.1}$ & $74.5{\scriptstyle\pm0.1}$ & $91.5{\scriptstyle\pm0.2}$ & $95.1{\scriptstyle\pm0.1}$ & $97.7{\scriptstyle\pm0.0}$ \\
\addlinespace[0.6ex]
\multicolumn{6}{l}{\textbf{\textit{Temporal Graphs}}} \\
DyRep & $41.6{\scriptstyle\pm3.8}$ & $26.9{\scriptstyle\pm4.4}$ & $47.8{\scriptstyle\pm4.0}$ & $58.4{\scriptstyle\pm3.2}$ & $72.5{\scriptstyle\pm2.0}$ \\
TGN & $59.1{\scriptstyle\pm3.2}$ & $44.5{\scriptstyle\pm4.0}$ & $68.9{\scriptstyle\pm3.1}$ & $77.6{\scriptstyle\pm2.1}$ & $86.1{\scriptstyle\pm1.0}$ \\
TCL & $72.2{\scriptstyle\pm1.4}$ & $60.0{\scriptstyle\pm1.8}$ & $81.5{\scriptstyle\pm1.2}$ & $87.6{\scriptstyle\pm0.8}$ & $93.5{\scriptstyle\pm0.5}$ \\
NAT & $78.8{\scriptstyle\pm4.0}$ & $69.7{\scriptstyle\pm5.1}$ & $85.5{\scriptstyle\pm3.8}$ & $90.7{\scriptstyle\pm2.7}$ & $95.4{\scriptstyle\pm1.3}$ \\
GraphMixer & $55.0{\scriptstyle\pm0.8}$ & $40.8{\scriptstyle\pm1.1}$ & $63.2{\scriptstyle\pm1.1}$ & $72.5{\scriptstyle\pm0.8}$ & $83.2{\scriptstyle\pm0.5}$ \\
TPNet & $42.5{\scriptstyle\pm0.9}$ & $27.5{\scriptstyle\pm0.7}$ & $51.4{\scriptstyle\pm1.2}$ & $61.6{\scriptstyle\pm1.2}$ & $71.8{\scriptstyle\pm0.9}$ \\
TNCN & $20.8{\scriptstyle\pm1.2}$ & $11.8{\scriptstyle\pm1.0}$ & $22.0{\scriptstyle\pm1.8}$ & $28.6{\scriptstyle\pm2.1}$ & $39.2{\scriptstyle\pm1.7}$ \\
\addlinespace[0.6ex]
\multicolumn{6}{l}{\textbf{\textit{\ourmodel}}} \\
\ourmodel-GCN & $\bm{90.4{\scriptstyle\pm0.4}}$ & $\bm{84.7{\scriptstyle\pm0.6}}$ & $\bm{95.6{\scriptstyle\pm0.2}}$ & $\bm{97.6{\scriptstyle\pm0.2}}$ & $\bm{99.0{\scriptstyle\pm0.1}}$ \\
\ourmodel-GAT & $89.1{\scriptstyle\pm0.0}$ & $82.4{\scriptstyle\pm0.1}$ & $\underline{95.2{\scriptstyle\pm0.0}}$ & $\underline{97.4{\scriptstyle\pm0.0}}$ & $\underline{98.9{\scriptstyle\pm0.0}}$ \\
\ourmodel-TFConv & $88.6{\scriptstyle\pm0.4}$ & $81.8{\scriptstyle\pm0.6}$ & $94.9{\scriptstyle\pm0.2}$ & $97.2{\scriptstyle\pm0.1}$ & $98.7{\scriptstyle\pm0.1}$ \\
\bottomrule
\end{tabular}
\end{table*}

\subsection{Significance testing}\label{app:significance}
To assess whether a difference exceeds three-seed variability, we serialize per-example predictions from both models at the validation-selected checkpoint on the \emph{identical} test split and run a \textbf{paired bootstrap over examples}: for each of $10{,}000$ resamples we draw the same examples for both models (class-stratified for the imbalanced fraud task), recompute the set-level metric for each, and take the gap, yielding a $95\%$ CI and a two-sided bootstrap $p$ per seed. We report this alongside the across-seed mean$\pm$std, because per-example (test-set) and across-seed (training) uncertainty answer different questions. For relationship prediction we additionally use a paired Wilcoxon signed-rank test on per-query reciprocal rank, since MRR decomposes per query whereas PR-AUC and Max-F1 are set-level ranking metrics that do not.

\Cref{tab:sig_fraud} applies this to the closest fraud comparison, RelGT$+$text versus \ourmodel-GAT on Clothing. The gap changes sign across seeds, only one of three seeds is individually significant on PR-AUC (seed~0, $p=0.028$) and none on Max-F1, and the across-seed gap ($+0.72\pm1.42$ PR-AUC, $+0.20\pm1.03$ Max-F1) lies well within noise. The seed-0 result is instructive: a per-example test on a \emph{single} training seed would report a ``significant'' advantage that does not replicate across seeds, precisely the false positive that across-seed variance corrects. We therefore report Clothing as a tie.

For relationship prediction, we apply the paired analysis to the matched freezing control (\cref{tab:matched_freeze}). For each seed we re-evaluate the validation-selected fine-tuned and frozen checkpoints on the identical $85{,}632$ test queries and candidate sets, and pair their per-query reciprocal ranks ($10{,}000$ bootstrap resamples; two-sided Wilcoxon signed-rank test). Fine-tuning wins on every seed by $9.4$--$10.6$ MRR points with tight confidence intervals (\cref{tab:sig_linkpred}). The re-evaluated means differ slightly from the logged ones in \cref{tab:matched_freeze} because neighborhood sampling at evaluation is stochastic and the evaluation batch size differs. We run these tests only for this matched pair and for the Clothing fraud comparison above.

\begin{table}[t]
\centering
\small
\setlength{\tabcolsep}{4pt}
\caption{Paired per-query test on Brightkite for the matched freezing control: MLM-pretrained encoder fine-tuned minus the same encoder frozen. MRR is in points ($\times100$), re-evaluated on the identical $85{,}632$ test queries; the $95\%$ CI is from $10{,}000$ bootstrap resamples over queries and $p$ from a two-sided Wilcoxon signed-rank test on reciprocal ranks.}
\label{tab:sig_linkpred}
\begin{tabular}{lcccc}
\toprule
\textbf{Seed} & \textbf{Fine-tuned} & \textbf{Frozen} & $\Delta$ \textbf{[95\% CI]} & \textbf{$p$} \\
\midrule
0 & $88.37$ & $79.00$ & $+9.36\ [+9.17,\,+9.56]$ & $<10^{-16}$ \\
1 & $88.46$ & $77.86$ & $+10.60\ [+10.39,\,+10.80]$ & $<10^{-16}$ \\
2 & $88.81$ & $79.12$ & $+9.69\ [+9.50,\,+9.88]$ & $<10^{-16}$ \\
\midrule
mean$\pm$std & & & $+9.88\pm0.64$ & \\
\bottomrule
\end{tabular}
\end{table}

\paragraph{Amazon product-overlap baseline.} Because Amazon relations are defined by co-reviewing at least three products while product IDs also appear in the input sequences, we evaluate a training-free baseline that uses only the model inputs. For each positive test relation and its $100$ negative candidates, we take the sets of full product IDs (ASINs) observed in both users' sequences and score a candidate by the size of their intersection, using mid-ranks for ties. \Cref{tab:amazon_overlap} reports the result; overlap rate is the share of positive pairs with at least one shared observed product. The shortcut is strongest on Movies, yet \ourmodel exceeds the baseline on every category. Masking the product IDs that define the relations would remove the shortcut directly; we leave it to future work.

\begin{table}[t]
\centering
\small
\setlength{\tabcolsep}{5pt}
\caption{Amazon observed-product overlap baseline (training-free) versus \ourmodel. Metrics in points; \ourmodel range over its GNN variants in \cref{tab:link_pred_full}.}
\label{tab:amazon_overlap}
\begin{tabular}{lccccc}
\toprule
\textbf{Dataset} & \textbf{Overlap rate} & \textbf{MRR} & \textbf{Hits@1} & \textbf{Hits@5} & \textbf{\ourmodel MRR} \\
\midrule
Movies & $71.6\%$ & $61.52$ & $52.2$ & $69.3$ & $88.6$--$90.4$ \\
Electronics & $43.0\%$ & $35.35$ & $26.5$ & $42.3$ & $75.1$--$80.2$ \\
Clothing & $29.8\%$ & $25.11$ & $18.9$ & $28.6$ & $52.3$--$63.7$ \\
\bottomrule
\end{tabular}
\end{table}

\begin{table}[t]
\centering
\small
\setlength{\tabcolsep}{4pt}
\caption{Paired per-user bootstrap on Amazon-Clothing, RelGT$+$text minus \ourmodel-GAT ($10{,}000$ class-stratified resamples over the shared test users, per seed). Gaps are in points ($\times100$) with $95\%$ CI and two-sided bootstrap $p$. The gap changes sign across seeds and only seed~0 (PR-AUC) is individually significant; the across-seed gap is within noise, so the two methods are tied on Clothing.}
\label{tab:sig_fraud}
\begin{tabular}{lcccc}
\toprule
 & \multicolumn{2}{c}{\textbf{PR-AUC gap}} & \multicolumn{2}{c}{\textbf{Max-F1 gap}} \\
\cmidrule(lr){2-3}\cmidrule(lr){4-5}
\textbf{Seed} & $\Delta$ \textbf{[95\% CI]} & \textbf{$p$} & $\Delta$ \textbf{[95\% CI]} & \textbf{$p$} \\
\midrule
0 & $+2.23\ [+0.28,\,+4.23]$ & $0.028$ & $+1.04\ [-1.41,\,+4.04]$ & $0.357$ \\
1 & $-0.59\ [-3.15,\,+1.91]$ & $0.627$ & $-0.95\ [-4.45,\,+2.27]$ & $0.527$ \\
2 & $+0.54\ [-1.82,\,+2.86]$ & $0.657$ & $+0.52\ [-3.24,\,+4.07]$ & $0.852$ \\
\midrule
mean$\pm$std & \multicolumn{2}{c}{$+0.72\pm1.42$} & \multicolumn{2}{c}{$+0.20\pm1.03$} \\
\bottomrule
\end{tabular}
\end{table}

\section{Temporal Graph Diagnostic}\label{sec:temporal_diag}

Temporal Graph methods \citep{luo2022neighborhood,yu_towards_2023, pmlr-v235-gravina24a} are popular methodologies for modeling user behavior that consists of interaction events. Their formulation focuses on modeling the dynamic evolution of edges (and nodes) over time by denoting the graph as sequences of timestamped edges \citep{luo2022neighborhood,zhang2406efficient,yu2023towards,gravina2024long}. This creates a mismatch when we want to apply these techniques to model user behavior that contains both dynamic personal user events and static user-to-user relationships, as their formulation does not support static relational information. Therefore, to apply temporal graph models, we are required to convert the original problem into an interaction event stream formulation.

For the relationship prediction task, each user's individual events are converted into timestamped edges between the user node and event nodes, and the static relationship edges must then be placed into this stream. Because friendship in Brightkite and Gowalla carries no time, these edges are assigned synthetic timestamps drawn at random from the overall span of observed event times; for the Amazon co-review graph, in contrast, the relation edges carry the real timestamps provided with the data. For the fraud detection task, each review becomes a timestamped user-item edge, static co-review edges must likewise be assigned timestamps to enter the stream, review text is attached as per-edge message features, ratings are added as edge features, and nodes are initialized from pooled Sentence-BERT embeddings. The results of these experiments serve as diagnostic rather than primary baseline comparisons: they test whether this conversion to a temporal-graph formulation is a viable substitute for the native formulation (Definition \ref{def:problem}) of user behavior modeling with sequential user event and static user-to-user relations.

\begin{table*}[t]
\caption{Relationship prediction temporal-graph diagnostic. Static graph methods use the given static user-relation graph directly. Temporal graph methods require converting user histories and static relations into a timestamped edge stream. Higher is better.}
\centering
\small
\setlength{\tabcolsep}{4pt}
\label{tab:temporal_link_diag}
\resizebox{\textwidth}{!}{%
\begin{tabular}{@{}l*{10}{c}@{}}
\toprule
& \multicolumn{2}{c}{\textbf{Brightkite}} & \multicolumn{2}{c}{\textbf{Gowalla}} & \multicolumn{2}{c}{\textbf{Movies}} & \multicolumn{2}{c}{\textbf{Electronics}} & \multicolumn{2}{c}{\textbf{Clothing}}\\
\cmidrule(lr){2-3}\cmidrule(lr){4-5}\cmidrule(lr){6-7}\cmidrule(lr){8-9}\cmidrule(lr){10-11}
\textbf{Method} & \textbf{MRR} & \textbf{H@5} & \textbf{MRR} & \textbf{H@5} & \textbf{MRR} & \textbf{H@5} & \textbf{MRR} & \textbf{H@5} & \textbf{MRR} & \textbf{H@5}\\
\midrule
\multicolumn{11}{l}{\textbf{\textit{Static Graph Methods}}} \\
GCN & $67.4{\scriptstyle\pm1.9}$ & $82.3{\scriptstyle\pm1.3}$ & $77.8{\scriptstyle\pm0.6}$ & $89.4{\scriptstyle\pm0.3}$ & $76.4{\scriptstyle\pm0.5}$ & $89.6{\scriptstyle\pm0.5}$ & $48.9{\scriptstyle\pm3.9}$ & $66.6{\scriptstyle\pm3.3}$ & $17.3{\scriptstyle\pm1.5}$ & $23.3{\scriptstyle\pm1.9}$ \\
GAT & $65.4{\scriptstyle\pm1.6}$ & $80.1{\scriptstyle\pm1.0}$ & $78.1{\scriptstyle\pm0.6}$ & $89.3{\scriptstyle\pm0.3}$ & $76.6{\scriptstyle\pm0.7}$ & $90.8{\scriptstyle\pm0.6}$ & $44.7{\scriptstyle\pm0.7}$ & $62.5{\scriptstyle\pm0.7}$ & $16.2{\scriptstyle\pm2.3}$ & $18.0{\scriptstyle\pm2.1}$ \\
TFConv & $68.5{\scriptstyle\pm0.2}$ & $83.2{\scriptstyle\pm0.4}$ & $80.5{\scriptstyle\pm0.2}$ & $90.9{\scriptstyle\pm0.2}$ & $82.4{\scriptstyle\pm0.1}$ & $94.4{\scriptstyle\pm0.0}$ & $55.8{\scriptstyle\pm0.7}$ & $71.5{\scriptstyle\pm0.4}$ & $18.8{\scriptstyle\pm0.7}$ & $25.2{\scriptstyle\pm0.3}$ \\
\addlinespace[0.6ex]
\multicolumn{11}{l}{\textbf{\textit{Temporal Graph Methods After Conversion}}} \\
DyRep & $26.4{\scriptstyle\pm3.7}$ & $37.5{\scriptstyle\pm5.5}$ & $27.8{\scriptstyle\pm2.1}$ & $37.8{\scriptstyle\pm3.3}$ & $41.6{\scriptstyle\pm3.8}$ & $58.4{\scriptstyle\pm3.2}$ & $17.0{\scriptstyle\pm1.9}$ & $23.7{\scriptstyle\pm2.1}$ & $13.3{\scriptstyle\pm0.8}$ & $18.9{\scriptstyle\pm1.3}$ \\
TGN & $38.7{\scriptstyle\pm1.6}$ & $53.9{\scriptstyle\pm1.3}$ & $44.8{\scriptstyle\pm2.7}$ & $59.3{\scriptstyle\pm3.4}$ & $59.1{\scriptstyle\pm3.2}$ & $77.6{\scriptstyle\pm2.1}$ & $33.8{\scriptstyle\pm1.5}$ & $46.5{\scriptstyle\pm2.1}$ & $14.8{\scriptstyle\pm1.4}$ & $20.8{\scriptstyle\pm3.0}$ \\
TCL & $20.4{\scriptstyle\pm1.0}$ & $29.1{\scriptstyle\pm1.5}$ & $20.5{\scriptstyle\pm1.7}$ & $27.5{\scriptstyle\pm2.5}$ & $72.2{\scriptstyle\pm1.4}$ & $87.6{\scriptstyle\pm0.8}$ & $34.9{\scriptstyle\pm0.2}$ & $47.3{\scriptstyle\pm0.1}$ & $14.2{\scriptstyle\pm0.5}$ & $20.5{\scriptstyle\pm1.3}$ \\
NAT & $68.2{\scriptstyle\pm7.8}$ & $77.0{\scriptstyle\pm5.9}$ & $59.1{\scriptstyle\pm13.0}$ & $68.0{\scriptstyle\pm11.8}$ & $78.8{\scriptstyle\pm4.0}$ & $90.7{\scriptstyle\pm2.7}$ & $62.8{\scriptstyle\pm8.0}$ & $70.4{\scriptstyle\pm6.4}$ & $30.6{\scriptstyle\pm7.9}$ & $36.8{\scriptstyle\pm7.2}$ \\
GraphMixer & $20.6{\scriptstyle\pm0.5}$ & $31.0{\scriptstyle\pm0.4}$ & $24.9{\scriptstyle\pm0.5}$ & $35.5{\scriptstyle\pm0.7}$ & $55.0{\scriptstyle\pm0.8}$ & $72.5{\scriptstyle\pm0.8}$ & $23.1{\scriptstyle\pm0.5}$ & $34.3{\scriptstyle\pm0.5}$ & $12.9{\scriptstyle\pm0.3}$ & $17.7{\scriptstyle\pm0.4}$ \\
TPNet & $45.0{\scriptstyle\pm0.6}$ & $53.6{\scriptstyle\pm0.4}$ & $51.4{\scriptstyle\pm1.0}$ & $62.9{\scriptstyle\pm1.3}$ & $42.5{\scriptstyle\pm0.9}$ & $61.6{\scriptstyle\pm1.2}$ & $22.9{\scriptstyle\pm0.2}$ & $34.7{\scriptstyle\pm0.3}$ & $11.4{\scriptstyle\pm0.6}$ & $14.8{\scriptstyle\pm1.1}$ \\
TNCN & $53.8{\scriptstyle\pm0.5}$ & $65.9{\scriptstyle\pm1.0}$ & $51.8{\scriptstyle\pm0.7}$ & $64.0{\scriptstyle\pm0.8}$ & $20.8{\scriptstyle\pm1.2}$ & $28.6{\scriptstyle\pm2.1}$ & $45.4{\scriptstyle\pm1.1}$ & $55.8{\scriptstyle\pm0.9}$ & $31.7{\scriptstyle\pm1.8}$ & $39.6{\scriptstyle\pm2.1}$ \\
\bottomrule
\end{tabular}
}
\end{table*}

\begin{table*}[t]
\caption{Fraud detection temporal-graph diagnostic. Static graph methods use averaged Sentence-BERT review embeddings as node features. Temporal graph methods receive review text and ratings through the converted timestamped edge stream. Higher is better.}
\centering
\small
\setlength{\tabcolsep}{4pt}
\resizebox{0.75\textwidth}{!}{%
\label{tab:temporal_fraud_diag}
\begin{tabular}{lcccccc}
\toprule
 & \multicolumn{2}{c}{\textbf{Amazon-Movies}} & \multicolumn{2}{c}{\textbf{Amazon-Electronics}} & \multicolumn{2}{c}{\textbf{Amazon-Clothing}} \\
\cmidrule(lr){2-3}\cmidrule(lr){4-5}\cmidrule(lr){6-7}
\textbf{Method} & \textbf{Max F1} & \textbf{PR AUC} & \textbf{Max F1} & \textbf{PR AUC} & \textbf{Max F1} & \textbf{PR AUC} \\
\midrule
\multicolumn{7}{l}{\textbf{\textit{Static Graph Methods + Text Embedding}}} \\
GAT & $78.3\,{\scriptstyle \pm 1.0}$ & $85.9\,{\scriptstyle \pm 1.5}$ & $46.0\,{\scriptstyle \pm 1.5}$ & $44.6\,{\scriptstyle \pm 2.5}$ & $37.9\,{\scriptstyle \pm 1.5}$ & $35.8\,{\scriptstyle \pm 2.4}$ \\
GCN & $78.5\,{\scriptstyle \pm 1.0}$ & $86.0\,{\scriptstyle \pm 1.5}$ & $46.0\,{\scriptstyle \pm 1.3}$ & $45.0\,{\scriptstyle \pm 2.7}$ & $38.0\,{\scriptstyle \pm 1.6}$ & $35.9\,{\scriptstyle \pm 2.3}$ \\
TFConv & $78.8\,{\scriptstyle \pm 0.9}$ & $86.6\,{\scriptstyle \pm 1.2}$ & $46.7\,{\scriptstyle \pm 0.9}$ & $45.1\,{\scriptstyle \pm 2.2}$ & $37.8\,{\scriptstyle \pm 1.9}$ & $35.4\,{\scriptstyle \pm 1.7}$ \\
\addlinespace[0.8ex]
\multicolumn{7}{l}{\textbf{\textit{Temporal Graph Methods After Conversion}}} \\
DyRep & $74.7\,{\scriptstyle \pm 0.7}$ & $82.3\,{\scriptstyle \pm 0.4}$ & $46.5\,{\scriptstyle \pm 0.3}$ & $46.3\,{\scriptstyle \pm 1.0}$ & $34.0\,{\scriptstyle \pm 2.8}$ & $33.2\,{\scriptstyle \pm 3.4}$ \\
GraphMixer & $69.1\,{\scriptstyle \pm 0.7}$ & $74.4\,{\scriptstyle \pm 0.3}$ & $32.7\,{\scriptstyle \pm 1.2}$ & $26.0\,{\scriptstyle \pm 1.3}$ & $19.8\,{\scriptstyle \pm 3.0}$ & $15.6\,{\scriptstyle \pm 3.6}$ \\
NAT & $51.2\,{\scriptstyle \pm 0.9}$ & $40.9\,{\scriptstyle \pm 0.5}$ & $18.2\,{\scriptstyle \pm 0.1}$ & $10.6\,{\scriptstyle \pm 0.2}$ & $8.8\,{\scriptstyle \pm 0.1}$ & $4.1\,{\scriptstyle \pm 0.3}$ \\
TCL & $74.8\,{\scriptstyle \pm 0.6}$ & $82.5\,{\scriptstyle \pm 0.1}$ & $32.6\,{\scriptstyle \pm 17.5}$ & $45.5\,{\scriptstyle \pm 6.8}$ & $14.9\,{\scriptstyle \pm 14.8}$ & $27.5\,{\scriptstyle \pm 24.0}$ \\
TGN & $75.5\,{\scriptstyle \pm 0.6}$ & $82.9\,{\scriptstyle \pm 0.3}$ & $46.8\,{\scriptstyle \pm 0.7}$ & $46.9\,{\scriptstyle \pm 1.3}$ & $35.0\,{\scriptstyle \pm 2.5}$ & $33.5\,{\scriptstyle \pm 2.9}$ \\
TPNet & $72.3\,{\scriptstyle \pm 0.8}$ & $79.2\,{\scriptstyle \pm 0.5}$ & $40.1\,{\scriptstyle \pm 0.8}$ & $35.8\,{\scriptstyle \pm 1.0}$ & $20.2\,{\scriptstyle \pm 1.9}$ & $15.6\,{\scriptstyle \pm 0.9}$ \\
\bottomrule
\end{tabular}
}
\end{table*}

\Cref{tab:temporal_link_diag} shows that temporal graph methods with this conversion are inconsistent in the relationship prediction tasks. Some temporal models are competitive on individual Amazon categories, but they do not reliably outperform the simpler static graph baselines across datasets. \Cref{tab:temporal_fraud_diag} shows the same pattern for fraud detection: even with review text, ratings, and initialized node embeddings, temporal graph methods do not consistently improve over static graph + text baselines. These results do not imply that temporal graph methods are weak in their intended setting; rather, they show that forcing static relations and individual user event sequences into a timestamped edge stream introduces artificial modeling choices without producing a stronger representation for our tasks.

\section{Ablation Studies}
\label{sec:ablation}
\begin{figure*}
    \centering
    \includegraphics[width=1\linewidth]{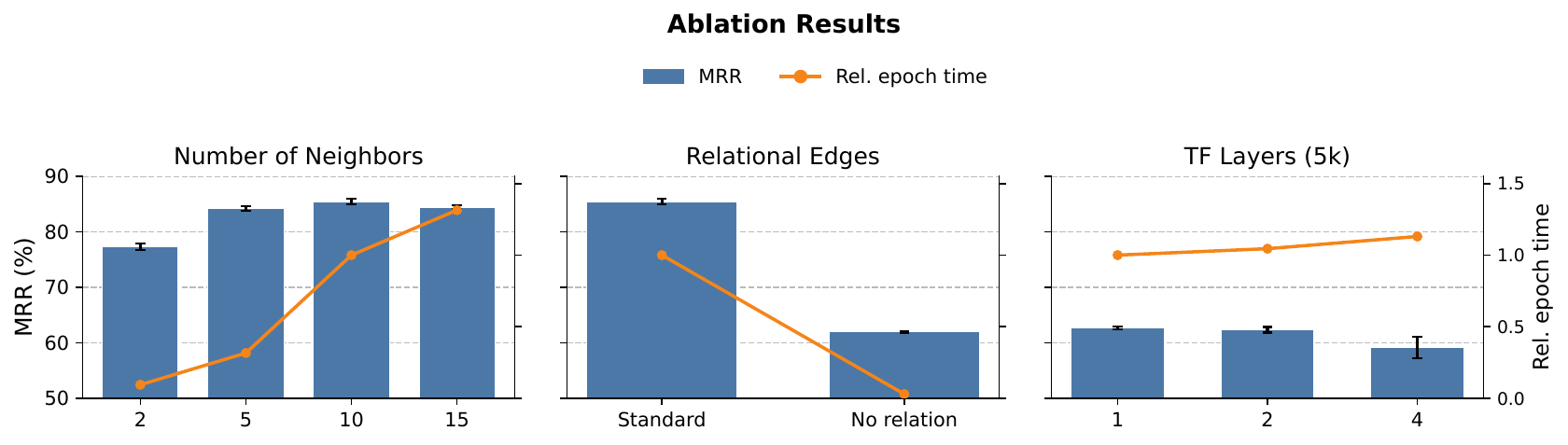}
    \caption{\textbf{Ablations on Brightkite.} Bars indicate test MRR (\%) and the orange line indicates relative epoch time (right axis) across (i) sampled neighbors, (ii) No-relation ablation, and (iii) sequence module depth. Runtime is normalized within each sweep to its reference setting.}
    \label{fig:ablate_fig}
\end{figure*}

We ablate three practical choices in \ourmodel on Brightkite dataset: neighbor fanout and sequence encoder depth, and the graph module. \Cref{fig:ablate_fig} reports MRR and relative epoch time; full values are in Appendix \ref{app:ablation}.

\paragraph{Number of Neighbors}
From the first panel of \Cref{fig:ablate_fig}, reducing the neighbor sampling size from 10 to 5 or 2 lowers the runtime substantially, with only a moderate drop in MRR. This shows that \ourmodel remains usable under tighter compute budgets, but also that BRIDGE benefits from access to nontrivial relational context. Overall, 10 or 15 neighbors offer the strongest performance,
while 2--5 neighbors provide a cheaper but weaker operating point when runtime is the primary constraint. Exact performance and epoch time are reported in Appendix \ref{app:ablation}.

\paragraph{No-relation ablation.}
We next evaluate a no-relation ablation that removes all edges from the message-passing graph while keeping the same sequence module, prediction objective, and end-to-end training procedure. This is stricter than setting the neighborhood fanout to zero: rather than modifying only the sampler, we remove the relational pathway itself while preserving the same train/validation/test supervision. 
The resulting comparison 
isolates the contribution of relational propagation beyond joint sequence modeling alone. Compared with the full BRIDGE model, the no-relation variant performs worse, showing that BRIDGE's gains do not come only from learning stronger sequence representations; they also depend on propagating information over the relational graph. On Brightkite, under the reduced ablation protocol (batch size $16$, $30$ epochs), removing relational message passing lowers MRR from $85.4\pm0.6$ to $62.0\pm0.2$ (\cref{tab:ablate_relational}). On Gowalla, we run the same ablation under the main-table setting instead: removing relational message passing lowers MRR from $87.9\pm0.2$ to $50.8\pm0.1$ (\cref{tab:ablate_relational}). Because the original Gowalla checkpoints were unavailable, this run uses the main-table configuration as recorded in our launch scripts, and its evaluation negatives are rebuilt from the public data release by keeping the first $100$ of each candidate set. This ablation isolates the value of the relational \emph{pathway} at inference, not the value of optimizing it end-to-end; the matched freezing control (\cref{tab:matched_freeze}) is what tests the training-strategy claim.

\paragraph{Transformer depth.}
The third panel of \Cref{fig:ablate_fig} examines the effect of sequence-module depth. Because deeper sequence encoders incur higher computational cost, we conduct this ablation on a 5{,}000-user induced subgraph of Brightkite dataset.

Increasing the number of Transformer layers from 1 to 2 yields nearly identical MRR, while increasing the depth from 2 to 4 slightly degrades performance. At the same time, relative epoch time increases modestly with depth. These results suggest that, on the ablation subgraph, a shallow sequence encoder already captures most of the useful sequential signal once combined with graph context. Accordingly, we adopt a 1-layer Transformer encoder as the default configuration in the main experiments, as it provides the best trade-off between performance and efficiency.

\paragraph{Parallelism and the cost of alternatives.} Because \ourmodel compresses each history into a single node vector before propagation, its graph phase is decoupled from sequence length (\cref{sec:complexity}) and its per-user Transformer encoders run in parallel across nodes. This contrasts with memory-based temporal graph models, which maintain and \emph{sequentially} update a per-node memory state as events arrive; routing high-frequency intransitive events (logins, clicks, check-ins) through such a memory turns the update into a throughput bottleneck. We report this as a structural scalability advantage of the static-relation formulation rather than a measured one; a matched throughput comparison is left to future work.

\paragraph{Cost of end-to-end training.} \Cref{tab:cost} compares the frozen and fine-tuned arms of the matched freezing control (\cref{tab:matched_freeze}) on the same hardware (one NVIDIA L40S, 48\,GB): architecture, batch size, initialization, and downstream settings are identical, and only encoder updating differs. Fine-tuning takes $2.2\times$ the time per epoch and $1.9\times$ the peak memory for its $10.6$-point MRR gain. We compare per-epoch time because early stopping ends the runs at different epochs. This accounting covers training only. Inference latency and the amortized benefit of reusing frozen embeddings across tasks are not measured: with on-the-fly encoding, both models take about $55$ minutes per validation pass, so the frozen pipeline's serving advantage would come from precomputing embeddings, which we do not time.

\subsection{Full ablation results}\label{app:ablation}
\Cref{tab:ablate_neighbors,tab:ablate_relational,tab:ablate_tfdepth} show the full results of ablation experiments. For each variant, we list the test MRR, per-epoch runtime (minutes), and runtime normalized by the reference configuration used in the main experiments (bolded).

\begin{table}[h]
\caption{\textbf{Sequence module depth.}
Sensitivity to the number of Transformer layers in the sequence encoder.
We report test MRR (mean$\pm$std over three seeds) and per-epoch runtime (minutes); \textbf{Rel. Epoch Time} is normalized to the two-layer reference used in this sweep.}
\centering
\small
\begin{tabular}{cccc}
\toprule
\# TF Layers & MRR & Epoch Time (min) & Rel. Epoch Time \\
\midrule
1 & $62.4\,{\scriptstyle \pm 0.4}$ & $43.8\,{\scriptstyle \pm 0.3}$ & 0.96 \\
2 & $62.4\,{\scriptstyle \pm 0.4}$ & $45.7\,{\scriptstyle \pm 0.1}$ & 1.00 \\
4 & $62.5\,{\scriptstyle \pm 0.3}$ & $49.5\,{\scriptstyle \pm 0.1}$ & 1.08 \\
\bottomrule
\end{tabular}
\label{tab:ablate_tfdepth}
\end{table}

\begin{table}[t]
\caption{Effect of neighborhood fanout on the full Brightkite friend-recommendation task. Mean and standard deviation are over three seeds.}
\centering
\begin{tabular}{cccc}
\toprule
\# Neighbors & MRR & Time (min) & Rel. Time \\
\midrule
2 & $77.3\,{\scriptstyle \pm 0.6}$ & $301.4\,{\scriptstyle \pm 120.4}$ & 0.09 \\
5 & $84.2\,{\scriptstyle \pm 0.4}$ & $1009.7\,{\scriptstyle \pm 96.1}$ & 0.31 \\
10 & $85.4\,{\scriptstyle \pm 0.6}$ & $3264.4\,{\scriptstyle \pm 770.9}$ & 1.00 \\
15 & $84.4\,{\scriptstyle \pm 0.3}$ & $4202.2\,{\scriptstyle \pm 609.0}$ & 1.29 \\
\bottomrule
\end{tabular}
\label{tab:ablate_neighbors}
\end{table}

\begin{table}[t]
\caption{Effect of relational message passing on friend recommendation. Mean and standard deviation are over three seeds. Brightkite uses the reduced ablation protocol (batch size 16, 30 epochs), so its full-model MRR ($85.4$) is below the main-table value ($92.2$, batch size 128) and the matched control ($87.4$, batch size 64, MLM initialization; \cref{tab:matched_freeze}). Gowalla uses the main-table setting; its ``with'' row is the main-table \ourmodel-GCN result.}
\centering
\begin{tabular}{llccc}
\toprule
Dataset & Variant & MRR & Time (min) & Rel. Time \\
\midrule
\multirow{2}{*}{Brightkite} & With relational MP& $85.4\,{\scriptstyle \pm 0.6}$ & $3264.4\,{\scriptstyle \pm 770.9}$ & 1.00 \\
 & No relational MP & $62.0\,{\scriptstyle \pm 0.2}$ & $93.9\,{\scriptstyle \pm 1.0}$ & 0.03 \\
\midrule
\multirow{2}{*}{Gowalla} & With relational MP & $87.9\,{\scriptstyle \pm 0.2}$ & -- & -- \\
 & No relational MP & $50.8\,{\scriptstyle \pm 0.1}$ & -- & -- \\
\bottomrule
\end{tabular}
\label{tab:ablate_relational}
\end{table}

\section{Limitations and Discussion}\label{app:limitations}

While \ourmodel\ provides a simple way to combine user event histories with static relation graphs, several limitations remain.

\subsection{Scope and Threats to Validity}\label{app:threats}
Our conclusions are empirical, and several factors could still affect them.
\begin{itemize}
\item \textbf{Confounders across datasets.} The matched freezing control (\cref{tab:matched_freeze}) isolates encoder updating on Brightkite only. Across our datasets, the task, label source, sequence modality (geohash tokens versus product IDs and text), and graph construction vary together, so ``the relation graph carries signal'' is one plausible explanation of where end-to-end training helps, not the only one.
\item \textbf{No within-dataset manipulation of the graph.} We do not drop, randomize, or shuffle edges, nor stratify queries by node degree, to vary graph signal within one dataset. These are the direct tests of the informativeness interpretation.
\item \textbf{Single-dataset matched control.} The frozen-versus-fine-tuned comparison from a shared checkpoint has not been repeated on Gowalla, Amazon, or fraud detection.
\item \textbf{Amazon shortcut.} The product-overlap baseline (\cref{tab:amazon_overlap}) bounds, but does not remove, the shortcut from product IDs that define Amazon relations; masking them is left to future work.
\item \textbf{Tuning budgets.} Relationship-prediction baselines largely follow settings from prior work rather than a per-method search of matched size (\cref{app:linkpred_hparams}), so differential tuning may contribute to cross-method gaps. The matched control is unaffected, since its two arms share all settings.
\item \textbf{Partial cost accounting.} We measure training time and memory (\cref{tab:cost}) but not inference latency or the amortized savings from reusing frozen embeddings.
\item \textbf{No theory.} We do not give a formal definition of graph informativeness or explain theoretically why it should govern the benefit of joint training.
\end{itemize}

\subsection{Use of Time Information}

\ourmodel\ preserves event order but does not explicitly encode absolute timestamps or inter-event time gaps.
For tasks where absolute time or time intervals are informative (e.g., periodicity or burstiness), incorporating relative/absolute time or time-aware positional encodings may yield further gains, which we leave to future work.

\subsection{Labels, Data, and Deployment}

The fraud detection labels are proxy labels derived from review helpfulness, so they should not be interpreted as verified ground truth fraud labels. The experiments use public datasets, which improves reproducibility but may not capture privacy constraints, fairness requirements, or distribution shifts that arise in deployed user behavior systems. Any deployment would require domain-specific validation and safeguards around sensitive user data.

\subsection{Computational Complexity}\label{sec:complexity}

A practical limitation of \ourmodel\ is computational cost when sequences are long or neighborhood fanout is large.
We summarize the dominant asymptotic terms to clarify the scalability trade-offs and motivate efficiency
improvements.

For each node $v_i\in\mathcal V$ with event sequence $S_i$ of length $M_i$, the sequence encoder produces token embeddings
$X_i\in\mathbb R^{M_i\times d}$.
We analyze the Transformer encoder used in our main experiments, 
and 
quantify the computation cost of the graph layer in \ourmodel-GCN.

\paragraph{Sequence encoding (Transformer).}
For a single Transformer layer on a length-$M_i$ sequence, the dominant cost comes from self-attention and the feed-forward blocks:
\begin{equation}
\mathcal O_{\text{TF}}(i) \;=\; \mathcal O\!\left(M_i^2 d \;+\; M_i d^2\right).
\end{equation}
The $M_i^2 d$ term corresponds to attention score computation and attention-weighted value aggregation, while $M_i d^2$ accounts for projections and MLPs.
Over all $N$ nodes, the total sequence-encoding cost is
\begin{equation}
\sum_{i=1}^N \mathcal O_{\text{TF}}(i)
\;=\;
\mathcal O\!\left(\sum_{i=1}^N (M_i^2 d + M_i d^2)\right).
\end{equation}
Although quadratic in $M_i$, Transformer encoding is highly parallelizable across tokens and across nodes.

\paragraph{Where the cost is paid in \ourmodel.}
The Transformer encoder is applied independently per node to obtain $\{X_i\}$. In \ourmodel-GCN (and other vector-feature GNN backbones), $X_i$ is compressed to a single vector before graph propagation; thus, the $M_i\times d$ feature maps need not be retained throughout message passing.

\paragraph{Computation cost in the graph module.}
To isolate the graph-phase overhead (holding the same batch size $B$, fanout $k$, and hidden size $d$),
\ourmodel-GCN compresses sequences before propagation and performs message passing on $d$-dimensional vectors, yielding
\begin{equation}
\mathcal O_{\text{GCN}} \;=\; \mathcal O(B\,k\,d^2).
\end{equation}
\ourmodel-GCN decouples graph propagation from sequence length,
making it scalable
for large graphs or long sequences.

\section{GenAI Usage Statement}\label{app:genai}
In this work, we used generative AI tools (large language model assistants) for language editing, for drafting revisions to the manuscript in response to reviewer feedback, and for coding assistance with experiment scripts (launch, data-preparation, and analysis scripts). We have reviewed all AI-assisted work: the authors revised all AI-drafted text and verified every reported number against the original result files. We take responsibility for the final content of this work, including text, claims, or artifacts produced with the aid of generative AI.

\end{document}